\documentclass{article}
\usepackage{graphicx} 
\usepackage{amsmath}
\usepackage{float} 
\usepackage{adjustbox}
\usepackage{url}

\usepackage{makecell} 
\usepackage{booktabs} 

\usepackage{cite}
\usepackage{pdflscape}
\usepackage{geometry}
\usepackage{comment}
\usepackage{caption} 
\usepackage{hyperref} 

\title{Revisiting Gradient Descent: A Dual-Weight Method for Improved Learning}

\author{XI WANG}
\date{\today}

\begin{document}

\maketitle

\begin{abstract}
    ****
\end{abstract}

\newpage
\tableofcontents

\newpage
\section{Introduction}
In neural networks, the weight vector \( W \) of a neuron plays a crucial role in transforming input features into outputs. While representing synaptic weights of postsynaptic neurons from presynaptic neurons, \( W \) can also be viewed as the neuron's encoding of the target concept it aims to represent. However, defining a target concept independently from other concepts often results in insufficient representation; rather, effective learning necessitates contrasting the target with non-targets. For instance, to accurately define a "dog," it is essential not only to understand the characteristics of dogs but also to distinguish them from non-dog entities. Without this contrast, differentiation remains incomplete.

Similarly, when a neuron learns, it should capture the differences between the features of the target class (hereafter termed positive examples) and those of non-target classes (negative examples). This suggests that the weight vector \( W \) should represent the contrast between the target features (\( W_1 \)) and the non-target features (\( W_2 \)), effectively modeling \( W = W_1 - W_2 \). Traditional methods, which store and update this contrast within a single weight vector \( W \) using gradient descent, as described by LeCun et al. (1989)~\cite{lecun1989backpropagation}, may not accurately capture this distinction.

In this study, we propose an alternative to the conventional gradient descent approach by decomposing the weight vector into two separate components, \( W_1 \) and \( W_2 \), which are updated independently. By explicitly modeling the contrast between target and non-target features, our method adjusts weights more accurately during training. Experimental results demonstrate that models trained using the formulation, \( (W_1 - W_2)X + \text{bias} \),  generalize better than those trained with the traditional \( WX + \text{bias} \) approach.

While our method introduces a slight increase in computational cost during training due to the separate updates of \( W_1 \) and \( W_2 \), the overall computational complexity remains equivalent to the traditional method in terms of Big-O notation. Moreover, once training is completed, we can directly compute and store \( W = W_1 - W_2 \), ensuring that the computational cost during the inference phase is identical to that of the conventional approach.

In summary, this research advocates for replacing the traditional single-weight gradient descent method with the \( W_1 - W_2 \) approach, offering a more accurate and effective framework for training neural networks.

\section{Related Work}
The idea that neurons should learn by contrasting positive and negative features has its roots in the broader field of contrastive learning~\cite{hadsell2006dimensionality,chen2020simclr,he2020moco}. Classical contrastive learning frameworks typically focus on comparing different input samples (e.g., pairs or triplets) to encourage representations that capture meaningful distinctions between similar and dissimilar instances. In contrast, our approach shifts the focus from sample-level comparisons to neuron-level weight contrasts. Instead of promoting discrimination by directly comparing input samples, we advocate comparing the weights a neuron learns under different conditions—i.e., how well it encodes target features relative to non-target features. This weight-centric perspective provides a more fine-grained mechanism for achieving contrastive learning within a neuron’s parameters. We will further elaborate on how our method differs from traditional contrastive learning methodologies in Appendix [A].

Our proposed update strategy also bears some conceptual resemblance to L2 regularization~\cite{goodfellow2016deep,hoerl1970ridge}, which is commonly employed to curb overfitting by penalizing large weight magnitudes. However, unlike L2 regularization—which uniformly applies a penalty to all weights regardless of their role or importance—our approach functions as a neuron-specific, adaptive form of regularization. This dynamic procedure allows each neuron to determine which time and how much to adjust its weights, thereby avoiding the indiscriminate suppression of valuable feature representations. By allowing neurons to dynamically control their own weight adjustments, our method avoids the potential downsides of blanket regularization, such as suppressing useful feature representations. Subsequent sections, including those dedicated to the comparison with L2 regularization and our experimental evaluation, will provide a more detailed discussion of these differences.

\section{Methodology}

Understanding how neurons encode information is crucial for improving neural network training methods. Traditional gradient descent methods update the synaptic weight vector \( W \) without explicitly distinguishing between target and non-target features. In this study, we propose a novel approach that decomposes the weight vector into two separate components, \( W_1 \) and \( W_2 \), to more accurately model the contrast between these features.

\subsection{Traditional gradient descent}

In the following, we focus on the computation performed by a single neuron. In traditional neural networks, the output of a neuron \( a \) is computed through a two-step process:
\begin{enumerate}
\item  Linear Combination:
   \begin{align}
   z = W X + b.
   \end{align}
   Here, \( W \) is the weight vector. \( X \) is the input vector. \( b \) is the bias term.
\item Activation:
   \begin{align}
   a = f(z),
   \end{align}
   where  \( f \) is the activation function applied to \( z \).
\end{enumerate}

Our objective is to adjust the weights \( W \) to minimize a loss function \( E \). An example of $E$ is the mean squared error, which measures the discrepancy between the network's predictions and the actual targets. To minimize \( E \), we use gradient descent to update the weights in the direction that reduces the loss:
\begin{align}
    \Delta W = -\eta \frac{\partial E}{\partial W},
\end{align}
where \( \Delta W \) is the change in weights. \( \eta \) is the learning rate, controlling the step size of each update. \( \frac{\partial E}{\partial W} \) is the gradient of the loss function with respect to the weights.

Using the chain rule, the gradient \( \frac{\partial E}{\partial W} \) can be written as:
\begin{align}
\frac{\partial E}{\partial W} 
&= \frac{\partial E}{\partial a} \cdot \frac{\partial a}{\partial z} \cdot \frac{\partial z}{\partial W} \nonumber\\
&= \frac{\partial E}{\partial a} \cdot f'(z) \cdot X
\end{align}
Hence, the weight update rule becomes:
\begin{align}  
 W_{\text{new}} = W_{\text{old}} + \Delta W,
\end{align}
where
\begin{align}
\Delta W = -\eta \cdot \frac{\partial E}{\partial a} \cdot f'(z) \cdot X.
\end{align}

\vskip 1em
\textbf{The case with ReLU activation function:} When the activation function \( f \) is the Rectified Linear Unit (ReLU), defined as \( f(z) = \max(0, z) \), its derivative \( f'(z) \) simplifies significantly:
\begin{align}
f'(z) =
\begin{cases}
1, & \text{if } z > 0 \\
0, & \text{if } z \leq 0
\end{cases}
\end{align}
Substituting \( f'(z) \) into the weight update rule:
\begin{align}  
  \Delta W = 
  \begin{cases}
      -\eta \cdot \frac{\partial E}{\partial a} \cdot X, & \text{if } z > 0 \\
      0, & \text{if } z \leq 0
  \end{cases}
\end{align}
This means that only neurons with a positive pre-activation value \( z \) (i.e., active neurons) will have their weights updated. Inactive neurons (where \( z \leq 0 \)) do not contribute to the weight update.

\subsubsection{Cumulative weight updates over iterations}

Considering multiple training iterations, the cumulative weight becomes:

\begin{align} 
    W = W_{\text{init}} - \eta \sum_{i=1}^{n} \text{grad}_i \cdot X_i,
\end{align}
where the gradient is defined as
\begin{align}
    \text{grad}_i \equiv \frac{\partial E}{\partial a} f'(z_i)
\end{align}
while \( W_{\text{init}} \) is the initial weight vector, and \( X_i \) and \( z_i \) represent an input vector at iteration \( i \) and weighted linear sum of its elements, respectively. After numerous updates, the influence of \( W_{\text{init}} \) diminishes, allowing us to focus on the aggregated updates. Since \( W_{\text{init}} \) is negligibly smaller than the second term, here we remove it in the following equations. 

Since gradients \( \text{grad}_i \) can be positive or negative, we can partition the summation based on the sign of the gradients:
\begin{align} 
    W = \eta \left( \sum_{\text{grad}_i < 0} (- \text{grad}_i) X_i \right) - \eta \left( \sum_{\text{grad}_i > 0} \text{grad}_i X_i \right).
    \label{eq:weight_decomposition_traditional}
\end{align}
This formulation clarifies that the weight vector \( W \) is effectively the difference between two aggregated terms: one associated with negative gradients (reinforcing the neuron's response to positive examples) and one with positive gradients (reducing the response to negative examples).

\subsection{Proposed weight decomposition and its implementation}

Recognizing the inherent structure with respect to the positive or negative gradient associated with each input in the traditional weight update rule (Eq.~\ref{eq:weight_decomposition_traditional}), we propose to extend it to perform the updates separately. Namely, we want the weight vector to be explicitly composed of the following two components:
\begin{align}
    W = W_1 - W_2,
\end{align}
where
\begin{align}
    W_1 &= \frac{\eta}{\sum_{\text{grad}_i < 0} (-\text{grad}_i)} \sum_{\text{grad}_i < 0} (-\text{grad}_i) X_i, \\
    W_2 &= \frac{\eta}{\sum_{\text{grad}_i > 0} \text{grad}_i} \sum_{\text{grad}_i > 0} \text{grad}_i X_i.
\end{align}
Here, 
\( W_1 \) represents the weighted average of the input vectors associated with negative gradients (i.e., when the neuron should increase its activation), and \( W_2 \) represents the weighted average associated with positive gradients (i.e., when the neuron should decrease its activation). By updating \( W_1 \) and \( W_2 \) separately, the model would more accurately capture the contrast between target and non-target features.

In practice, calculating the exact weighted average of all samples can be computationally demanding. Therefore, we use the following moving average to replace the average in the above formula of our method (See Appendix \ref{sec:convergence_proof} and \ref{sec:convergence_proof2}
 for the proof of equivalence). Our weight update formulas are now given as follows:
\begin{align}
\text{If } \text{grad}_i < 0,& \quad W_1^{\text{new}} = W_1^{\text{old}} \times (1 - \eta \cdot |\text{grad}_i|) + X_i \cdot \eta \cdot |\text{grad}_i|
\\
\text{If } \text{grad}_i > 0,& \quad W_2^{\text{new}} = W_2^{\text{old}} \times (1 - \eta \cdot |\text{grad}_i|) + X_i \cdot \eta \cdot |\text{grad}_i|
\\
\text{If } \text{grad}_i = 0,& \quad W_1^{\text{new}}=W_1^{\text{old}}, W_2^{\text{new}}=W_2^{\text{old}}
\end{align}
This moving average allows for efficient updates while ensuring that the core principle of averaging across all samples is retained.

\subsection{Biological Inspiration: Excitatory–Inhibitory Interplay}

In biological neural circuits, \textbf{excitatory neurons} increase their targets’ firing rates, while \textbf{inhibitory neurons} suppress or modulate that activity. This balance of excitation and inhibition helps the brain avoid runaway activity, maintain stability, and sharpen selectivity. A key mechanism in this balance is \textbf{lateral inhibition}, where inhibitory neurons learn to “push back” whenever a neighboring excitatory neuron’s response is higher than expected.

\begin{figure}[h]
    \centering
    \includegraphics[width=0.5\linewidth]{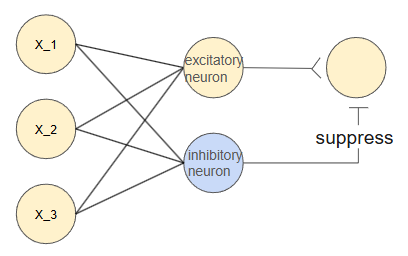}
    \caption{Both the excitatory neuron (top) and the inhibitory neuron (bottom) receive the same inputs \(\{x_1, x_2, x_3\}\). The excitatory neuron uses \(W_1\) as its weight vector, while the inhibitory neuron uses \(W_2\). Their combined outputs then feed into a final output neuron (right).}
    \label{fig:exc_inh_framework}

\end{figure}

We can draw an analogy between this process and our proposed decomposition of the weight vector into \(W_1\) (excitatory) and \(W_2\) (inhibitory) like Figure \ref{fig:exc_inh_framework}. At inference time, the neuron’s total output is given by
\[
W X = (W_1 - W_2) X,
\]
meaning that the inhibitory portion, \(W_2\), \textit{always} exerts some degree of suppressive influence on the net signal. However, learning for the inhibitory neuron only happens when the combined output \((W_1 - W_2) X\) is \textit{too large}—in other words, when there is a discrepancy between the neuron’s actual output and its desired (or “expected”) range. In that event, the inhibitory neuron’s weights are adjusted to be more similar to the current input \(X\), thereby increasing future suppression for that same input pattern.

This selective adjustment aligns well with how inhibitory neurons function in the brain: they monitor excitatory activity and learn to curb it if it strays above an adaptive threshold. By capturing this local interplay between excitation and inhibition, our weight decomposition naturally supports more stable and controlled responses, analogous to the self-regulating processes observed in biological neural circuits.

\subsection{Advantages of using mean over sum in weight updates}

Replacing the traditional summation in weight updates with a mean and decomposing the weight vector \( W \) into two separate components \( W_1 \) and \( W_2 \) offers several significant advantages:

\begin{itemize}
\item
\textbf{Mitigation of bias due to sample imbalance:} 
In datasets with imbalanced samples, traditional weight updates may become biased toward the majority feature, as the cumulative sum of gradients can disproportionately reflect the dominant samples. By calculating \( W_1 \) and \( W_2 \) separately for positive and negative gradients and using their means, the method addresses this potential bias. This separation ensures that both positive and negative features are equally represented in the weight updates, leading to a more balanced and fair model.

    \item 
\textbf{Elimination of residual weights in uninformative dimensions:} Traditional gradient descent updates only the dimensions corresponding to non-zero entries in the input vector \( X \). As a result, weights associated with zero-valued features remain unchanged, potentially leaving residual values in unimportant dimensions. These residual weights can cause unintended activations when the input distribution shifts, as they may respond to noise or irrelevant features. Our method synchronously updates all dimensions by incorporating a term \( -W \cdot | \text{grad} | \cdot \eta \), which gradually reduces the weights in unimportant dimensions toward zero when the neuron receives a non-zero gradient, but the corresponding features in \( X \) are frequently zero. This process enhances generalization by focusing the model on informative features and reducing the influence of irrelevant ones.

\item 
\textbf{Dynamic regularization based on feature importance:} By utilizing the magnitude of the gradient as a dynamic regularization parameter, the method adjusts the strength of regularization according to the impact of each output on the final result. In our definitions of \( W_1 \) and \( W_2 \), the gradient, \( \text{grad} \), determines the weight assigned to each input \( X \) during averaging. When \( \text{grad} = 0 \), the input \( X \) does not affect the weights, unlike uniform L1 or L2 regularization methods that apply regularization regardless of the neuron's activity. This adaptive regularization focuses learning on more significant samples, enhancing the model's ability to prioritize important information and improving overall performance.

\item 
\textbf{Enhanced generalization and focused learning:} The proposed approach ensures that the neuron's weights are concentrated on dimensions with high contributions to the output, improving the model's ability to generalize to new data.  By applying weight regularization only when a neuron is activated (i.e., $\text{grad}_i \neq 0$), the method effectively preserves the weights of neurons encoding low-frequency features. This mechanism prevents unnecessary penalization of inactive neurons while still reducing the weights of input dimensions that are less aligned with the neuron's activation frequency. Consequently, the model avoids overfitting to less significant dimensions and maintains sensitivity to low-frequency but crucial features, resulting in a more focused and adaptive representation. In contrast, the traditional sum-based method may fail to effectively reduce weights in unimportant dimensions, leading to a less focused model.

\item 
\textbf{Prevention of gradient explosion:} By computing \( W_1 \) and \( W_2 \) as weighted averages of input vectors, their values are inherently constrained within the range of the input data \( X \). Since \( W_1 \) and \( W_2 \) can be regarded as expected values of \( X \) over certain distributions, their magnitudes remain bounded as long as the distributions have finite mean. This limitation on the range of \( W = W_1 - W_2 \) prevents the weights from growing excessively large, thereby reducing the risk of gradient explosion during training and enhancing the stability of the learning process. The downside of this approach is that the outputs of the network are limited to a finite range. Hence, the method requires the appropriate scaling for the training values of the outputs. 

\end{itemize}


In summary, by replacing the traditional summation with a mean in weight updates and decomposing the weight vector into two components—$W_1$, which captures the neuron's target features, and 
$W_2$, which captures features not associated with the neuron—the proposed method introduces dynamic regularization based on gradient magnitude. This approach is expected to enhance the model's generalization ability, focus learning on significant features, and mitigate issues arising from sample imbalance and residual weights.

\subsection{Comparison with L2 regularization}

The method proposed in this study shares similarities with traditional regularization techniques like L2 regularization in terms of enhancing a model's generalization ability. However, there are critical differences that set our approach apart. This section aims to compare our method with L2 regularization, highlighting the unique advantages of our approach while avoiding redundant explanations.

L2 regularization adds a penalty proportional to the square of the magnitude of the weights, encouraging smaller weights and helping to prevent overfitting. The loss function with L2 regularization is defined as
\begin{align}
\text{Loss} = \frac{1}{2} \| y_{\text{pred}} - y_{\text{true}} \|^2 + \frac{\lambda}{2} \| W \|^2,
\end{align}
where \( \lambda \) is the regularization strength. The corresponding weight update equation, assuming the ReLU activation function and a learning rate \( \eta \), is
\begin{align}
W := W (1 - \eta \cdot \lambda) - \eta \cdot \text{grad} \cdot X.
\end{align}
There are key differences between our method and the advantages of our method, which are summarized as follows. 

\textbf{Separate updates for positive and negative features:} L2 Regularization updates all weights collectively without distinguishing between different types of features, while our Method splits the weight vector into \( W_1 \) and \( W_2 \), which are updated separately based on the sign of the gradient. This allows for more precise adjustments, particularly in datasets with imbalanced features, ensuring that less frequent features will be treated as equally important when updated.

\textbf{Dynamic regularization based on activation:} Moreover, L2 Regularization applies a uniform shrinkage to all weights during each update, regardless of whether the neuron is active or not. This means that even weights associated with neurons capturing rare but important features are penalized, potentially diminishing their influence. In contrast, our method replaces the fixed regularization parameter \( \lambda \) with the dynamic term \( \text{grad} \), effectively applying regularization only when the neuron is activated (i.e., when \( \text{grad} \neq 0 \)). This ensures that important weights corresponding to low-frequency but significant features are preserved, enhancing the model's expressiveness and generalization.

In summary, traditional L2 regularization methods penalize weights uniformly based on their magnitude, without accounting for their specific contributions to the neuron's output. In contrast, our proposed method evaluates the actual contribution of each feature to the neuron's output and separately updates weights for positive and negative features. This approach enables more precise and dynamic weight adjustments, effectively preserving important weights while reducing the influence of weights associated with dimensions that provide limited value for feature encoding, thereby mitigating the risk of overfitting on less relevant dimensions.








\section{Results}

\subsection{Experimental setup}
The aim of our experiments is to compare the learning capabilities of the proposed weight update method with the traditional gradient descent and L2 regularization methods under identical conditions. To better reveal the model’s learning ability, we designed scenarios that are prone to overfitting using a small sample size. Each method creates multiple models trained under different conditions (e.g., varying network sizes, training sample sizes, and datasets). We then compare models under the same conditions to evaluate which method performs better.

We included the following training methods in our experiments:
\begin{itemize}
    \item \textbf{Proposed method}

    \item \textbf{Proposed method with simple stabilizer}

    \item \textbf{Traditional gradient descent method} 

    \item \textbf{L2 regularization method} 
\end{itemize}

Note that our proposed method currently lacks a matching optimizer, such as Adam (Kingma and Ba, 2014)~\cite{kingma2014adam}, and batch training technique, so we can only train by randomly learning one sample at a time. This introduces excessive randomness and instability,  as highlighted in prior studies on the importance of batch training for stabilizing learning (Goodfellow et al., 2016)~\cite{goodfellow2016deep}. Including a simple stabilizer allows us to better demonstrate the potential of our method once appropriate optimizers and batch training techniques are developed. We thus constructed a stabilizer for learning and added it for comparison. The formula of this stabilizer is written in the appendix B. The traditional gradient descent method serves as the baseline for comparison. We also tested L2 regularization with penalty weights of 0.1, and 0.01 to assess its impact under different regularization strengths.

All experiments were conducted under the following conditions regarding the network, training, and the data:
\begin{itemize}
    \item \textbf{Network architectures}
    \begin{itemize}
        \item \textbf{2 Layers Network}: Input dimension $\rightarrow$ 20 neurons $\rightarrow$ Output dimension.
        \item \textbf{3 Layers Network}: Input dimension $\rightarrow$ 64 neurons $\rightarrow$ 32 neurons $\rightarrow$ Output dimension.
    \end{itemize}

    \item \textbf{Training}
    \begin{itemize}
        \item \textbf{Method}: The bias values for all methods were updated using the basic stochastic gradient descent method. Only the update methods for the weight matrix $W$ differed.

        \item \textbf{Initialization}: All models were initialized with the same parameter values to ensure fairness.
        
        \item \textbf{Iterations}: Each model was trained for an equal number of iterations (60,000).

    \end{itemize}

    \item \textbf{Data}
    \begin{itemize}
        \item \textbf{Identitical samples}: The same training samples were used across all methods.
        
        \item \textbf{Sample sizes}: Models were trained on small-scale datasets consisting of 100, 500, and 1,000 samples to create scenarios prone to overfitting.
    \end{itemize}
\end{itemize}

To comprehensively evaluate the methods, we conducted experiments on both classification and regression tasks:
\begin{itemize}
    \item \textbf{Regression Tasks} California Housing~\cite{scikit-learn} dataset and Wine Quality~\cite{cortez2009modeling} dataset. Performance was assessed based on the loss on the test datasets.

\item \textbf{Classification Tasks} MNIST~\cite{lecun1998gradient}, Fashion-MNIST~\cite{xiao2017fashion}, and CIFAR-10~\cite{krizhevsky2009learning} datasets. Model performance was measured by accuracy on the test datasets.
\end{itemize}

Some datasets were normalized before training by subtracting the mean and dividing by the standard deviation. Additionally, since every neuron's output value in our proposed method has a theoretical upper limit, the method may not function properly if the target \( y \) values in regression tasks exceed this limit. To address this, we applied a scaling factor by multiplying the normalized \( y \) values by 0.1, keeping them within the acceptable range without compromising the fairness of the experiments.

To further evaluate the generalization capabilities of the trained models, we simulated a distribution shift between the training and test samples—a common scenario in real-world applications. After training, the models were tested not only on the standard test samples but also on samples with added noise (\( \text{noise} \sim \mathcal{N}(0, 0.3) \)) to assess their robustness.


\subsection{Experimental results}
We conducted experiments across the five datasets and two model configurations to compare five training methods. The outcomes for both regression and classification tasks are summarized in Table \ref{originaltable_regression} and \ref{originaltable_classification}, respectively. For regression datasets, reported values are the mean test loss multiplied by 10,000. For classification datasets, values are the mean test accuracy multiplied by 100. All reported values correspond to test performance only; training performance is not recorded.

\begin{table}[!th]
    \centering
    \caption{Average Losses in Regression Tasks}
    \resizebox{\textwidth}{!}{
    \begin{tabular}{c|c|c|c|c|c|c|c|c|c}
    \toprule
    \makecell{} & \makecell{Dataset} & \makecell{Layer} & \makecell{Method} & 
    \makecell{Normal \\ 100 samp} & \makecell{Noise \\ 100 samp} & 
    \makecell{Normal \\ 500 samp} & \makecell{Noise \\ 500 samp} & 
    \makecell{Normal \\ 1000 samp} & \makecell{Noise \\ 1000 samp} \\ \midrule
        0 & wine & 2 & our & 102.0 & 105.7 & 77.8 & 80.5 & 74.4 & 76.9 \\ 
        1 & wine & 2 & our+stabilizer & \underline{92.0} & \underline{96.0} & \underline{70.8} & \underline{72.9} & \underline{68.5} & \underline{70.3} \\ 
        2 & wine & 2 & gradient descent & 260.2 & 320.4 & 76.7 & 80.8 & 73.3 & 76.1 \\ 
        3 & wine & 2 & L2 0.01 & 115.3 & 123.5 & 79.6 & 81.4 & 75.0 & 77.7 \\ 
        4 & wine & 2 & L2 0.1 & 100.1 & 100.1 & 100.2 & 100.2 & 99.3 & 99.3 \\ \hline
        5 & wine & 3 & our & 117.7 & 126.6 & 76.4 & 79.3 & 76.6 & 79.5 \\ 
        6 & wine & 3 & our+stabilizer & 102.7 & 107.5 & \underline{69.9} & \underline{72.8} & \underline{67.8} & \underline{70.6} \\ 
        7 & wine & 3 & gradient descent & 653.4 & 791.6 & 108.5 & 118.1 & 91.0 & 101.5 \\ 
        8 & wine & 3 & L2 0.01 & 232.9 & 264.6 & 75.8 & 79.8 & 73.4 & 76.0 \\ 
        9 & wine & 3 & L2 0.1 & \underline{100.1} & \underline{100.1} & 100.2 & 100.2 & 99.3 & 99.3 \\ \hline
        10 & house & 2 & our & 47.6 & 57.6 & 47.7 & 56.2 & 49.5 & 58.3 \\ 
        11 & house & 2 & our+stabilizer & \underline{43.2} & \underline{51.7} & \underline{39.6} & \underline{47.7} & \underline{38.1} & \underline{46.1} \\ 
        12 & house & 2 & gradient descent & 74.4 & 147.9 & 65.7 & 158.9 & 53.4 & 83.4 \\ 
        13 & house & 2 & L2 0.01 & 49.5 & 69.8 & 49.0 & 73.8 & 54.5 & 63.0 \\ 
        14 & house & 2 & L2 0.1 & 100.0 & 100.0 & 100.0 & 100.0 & 100.0 & 100.0 \\ \hline
        15 & house & 3 & our & 49.5 & 63.0 & 38.7 & 54.5 & 44.6 & 57.3 \\ 
        16 & house & 3 & our+stabilizer & \underline{43.1} & \underline{51.8} & \underline{37.5} & \underline{46.1} & \underline{36.3} & \underline{45.1} \\ 
        17 & house & 3 & gradient descent & 168.5 & 286.0 & 50.0 & 107.9 & 46.8 & 83.5 \\ 
        18 & house & 3 & L2 0.01 & 72.2 & 104.4 & 38.4 & 57.6 & 37.6 & 53.7 \\ 
        19 & house & 3 & L2 0.1 & 99.3 & 99.3 & 100.0 & 100.0 & 100.0 & 100.0 \\

    \bottomrule
    \end{tabular}}
    \caption*{
    Dataset: The dataset is used to train this model. Layer: Indicates the model architecture (e.g., the number of layers). Method: Specifies the training method used (e.g., gradient descent, our proposed method, or L2 regularization methods). Columns such as "Normal 100 samp" or "Noise 100 samp": Each column represents test performance under a specific training set size and testing condition. 
    }
    \label{originaltable_regression}
\end{table}

\begin{table}[!th]
    \centering
    \caption{Average Accuracy (Percentage) in Classification Tasks}
    \resizebox{\textwidth}{!}{ 
    \begin{tabular}{c|c|c|c|c|c|c|c|c|c}
    \toprule
    \makecell{} & \makecell{Dataset} & \makecell{Layer} & \makecell{Method} & 
    \makecell{Normal \\ 100 samp} & \makecell{Noise \\ 100 samp} & 
    \makecell{Normal \\ 500 samp} & \makecell{Noise \\ 500 samp} & 
    \makecell{Normal \\ 1000 samp} & \makecell{Noise \\ 1000 samp} \\ \midrule
        20 & mnist & 2 & our & 74.9 & 67.7 & \underline{85.6} & 80.6 & 87.5 & \underline{83.7} \\ 
        21 & mnist & 2 & our+stabilizer & 75.2 & 66.0 & \underline{85.6} & \underline{81.2} & 87.6 & 82.0 \\ 
        22 & mnist & 2 & gradient descent & 73.8 & 66.1 & 84.0 & 77.4 & 87.7 & 73.6 \\ 
        23 & mnist & 2 & L2 0.01 & 74.0 & 66.8 & 83.8 & 77.2 & 87.8 & 74.7 \\ 
        24 & mnist & 2 & L2 0.1 & \underline{75.4} & \underline{69.5} & 84.8 & 80.4 & \underline{87.9} & 83.1 \\ \hline
        25 & mnist & 3 & our & 74.7 & 68.2 & 85.7 & 79.2 & 88.6 & 82.3 \\ 
        26 & mnist & 3 & our+stabilizer & 73.3 & 62.2 & \underline{85.9} & 78.0 & \underline{89.0} & 82.2 \\ 
        27 & mnist & 3 & gradient descent & 72.9 & 64.8 & 85.1 & 76.3 & 88.4 & 78.6 \\ 
        28 & mnist & 3 & L2 0.01 & 73.8 & 65.8 & 85.1 & 76.3 & 88.5 & 79.8 \\ 
        29 & mnist & 3 & L2 0.1 & \underline{75.8} & \underline{70.0} & 85.4 & \underline{80.0} & 88.7 & \underline{83.9} \\ \hline
        30 & fashion & 2 & our & 69.8 & 67.3 & 78.6 & 74.4 & 78.7 & \underline{76.5} \\ 
        31 & fashion & 2 & our+stabilizer & 69.7 & 65.8 & 78.0 & 71.0 & 78.8 & 75.9 \\ 
        32 & fashion & 2 & gradient descent & 68.7 & 66.4 & 78.8 & 72.6 & 79.5 & 73.1 \\ 
        33 & fashion & 2 & L2 0.01 & 69.4 & 66.9 & \underline{78.9} & 72.6 & \underline{80.5} & 74.8 \\ 
        34 & fashion & 2 & L2 0.1 & \underline{70.4} & \underline{69.2} & 78.3 & \underline{74.9} & 79.3 & 76.4 \\ \hline
        35 & fashion & 3 & our & 68.8 & 66.2 & \underline{79.0} & \underline{73.9} & 79.5 & 75.7 \\ 
        36 & fashion & 3 & our+stabilizer & 68.0 & 64.2 & 78.5 & 71.6 & \underline{79.8} & \underline{76.6} \\ 
        37 & fashion & 3 & gradient descent & 69.1 & 64.9 & 77.9 & 69.5 & 79.0 & 71.5 \\ 
        38 & fashion & 3 & L2 0.01 & 69.0 & 65.4 & 78.0 & 69.2 & 79.7 & 72.6 \\ 
        39 & fashion & 3 & L2 0.1 & \underline{69.8} & \underline{67.7} & 77.3 & \underline{73.9} & 79.4 & 75.9 \\ \hline
        40 & cifar10 & 2 & our & 24.2 & 22.7 & 23.4 & 21.8 & 30.5 & 28.9 \\ 
        41 & cifar10 & 2 & our+stabilizer & \underline{24.5} & \underline{23.3} & \underline{31.6} & \underline{30.2} & \underline{31.6} & \underline{30.9} \\ 
        42 & cifar10 & 2 & gradient descent & 23.1 & 21.8 & 27.9 & 25.0 & 29.3 & 26.7 \\ 
        43 & cifar10 & 2 & L2 0.01 & 22.8 & 21.9 & 29.2 & 26.6 & 30.6 & 28.0 \\ 
        44 & cifar10 & 2 & L2 0.1 & 24.3 & 23.0 & 30.2 & 28.5 & 30.6 & 29.0 \\ \hline
        45 & cifar10 & 3 & our & 24.6 & 23.3 & 31.6 & 28.9 & 32.1 & 30.3 \\ 
        46 & cifar10 & 3 & our+stabilizer & 23.8 & 22.9 & 30.5 & 28.0 & 33.4 & 31.5 \\ 
        47 & cifar10 & 3 & gradient descent & 23.3 & 21.7 & 31.6 & 29.1 & 33.7 & 30.9 \\ 
        48 & cifar10 & 3 & L2 0.01 & 24.9 & 23.2 & 30.5 & 28.2 & \underline{34.2} & \underline{31.7} \\ 
        49 & cifar10 & 3 & L2 0.1 & \underline{25.7} & \underline{24.5} & \underline{31.8} & \underline{29.2} & 33.7 & 31.2 \\ 

    \bottomrule
    \end{tabular}}
    \label{originaltable_classification}
\end{table}

We compared the performance of the methods for data sets of multiple sizes with or without noise. The columns in Table \ref{originaltable_regression} and \ref{originaltable_classification}, such as "Normal 100 samp" or "Noise 100 samp", represent test performance under a specific training set size and testing condition. For example, “Normal 100 samp” denotes that the model was trained on 100 samples, and tested on the standard (non-noise) test set. In contrast, “Noise 100 samp” still denotes that the model was trained on the same 100 non-noise samples, but tested on a noise-perturbed version of the test set. Similarly, “Normal 500 samp,” “Noise 500 samp,” and so forth follow the same logic.

Overall, on regression tasks, our proposed methods tend to produce lower test losses compared to both the ordinary gradient descent method and L2 regularization methods. On classification tasks, the differences among the methods were less pronounced.

\subsubsection{Comparison to gradient descent using relative measures}

To more intuitively show how different methods deviate from the base gradient descent result, we want to set the gradient descent method as baseline and compute the relative difference 
\[
\frac{\text{Baseline method} - \text{Method}}{\text{Baseline method}} \text{  (for regression tasks)}
\]
\[
\frac{\text{Method} - \text{Baseline method}}{\text{Baseline method}} \text{  (for classifications tasks)}
\]
for each result. Then, we averaged these relative differences over various training conditions for each dataset and method. 
Table \ref{avgtable_regression} and \ref{avgtable_classification} show these averaged relative results.

\begin{table}[!th]
    \centering
    \caption{Relative Differences in Average over Methods on Regression Tasks}
    \begin{tabular}{c|c|c|c|c}
    \toprule
    \makecell{} & \makecell{Dataset} & \makecell{Method} & 
    \makecell{Normal \\ Sample} & 
    \makecell{Noise \\ Sample} \\ \midrule

        0 & house & our & 28.1 & 52.4 \\ 
        1 & house & our+stabilizer & 38.7 & 60.8 \\ 
        2 & house & L2 0.01 & 26.1 & 46.1 \\ 
        3 & house & L2 0.1 & -57.7 & 17.1 \\ 
        4 & wine & our & 30.9 & 34.1 \\ 
        5 & wine & our+stabilizer & 37.4 & 40.4 \\ 
        6 & wine & L2 0.01 & 27.2 & 30.5 \\ 
        7 & wine & L2 0.1 & 13.1 & 19.8 \\ 
        
    \bottomrule
    \end{tabular}
    \label{avgtable_regression}
\end{table}

\begin{table}[!th]
    \centering
    \caption{Relative Differences in Average over Methods on Classification Tasks}
    \begin{tabular}{c|c|c|c|c}
    \toprule
    \makecell{} & \makecell{Dataset} & \makecell{Method} & 
    \makecell{Normal Sample \\ Improvement} & \makecell{Noise Sample \\ Improvement} \\ \midrule
        0 & cifar10 & our & -1.1 & 0.7 \\ 
        1 & cifar10 & our+stabilizer & 4.2 & 7.6 \\ 
        2 & cifar10 & L2 0.01 & 2.1 & 3.0 \\ 
        3 & cifar10 & L2 0.1 & 4.8 & 7.3 \\ 
        4 & fashion & our & 0.3 & 3.8 \\ 
        5 & fashion & our+stabilizer & -0.0 & 1.6 \\ 
        6 & fashion & L2 0.01 & 0.5 & 0.8 \\ 
        7 & fashion & L2 0.1 & 0.4 & 4.4 \\ 
        8 & mnist & our & 1.1 & 5.7 \\ 
        9 & mnist & our+stabilizer & 1.0 & 3.2 \\ 
        10 & mnist & L2 0.01 & 0.2 & 0.9 \\ 
        11 & mnist & L2 0.1 & 1.3 & 6.9 \\ 
        
    \bottomrule
    \end{tabular}
    \label{avgtable_classification}
\end{table}



\subsubsection{Averaeg performance for regression and classification tasks}
To summarize the results more clearly, we further averaged different datasets from Table \ref{avgtable_regression} and \ref{avgtable_classification}, and plotted the average deviations from the baseline for each task. Figures \ref{fig:Regression_Classification} highlight these differences:

\begin{figure}[!t]
    \centering
    \includegraphics[width=0.75\linewidth]{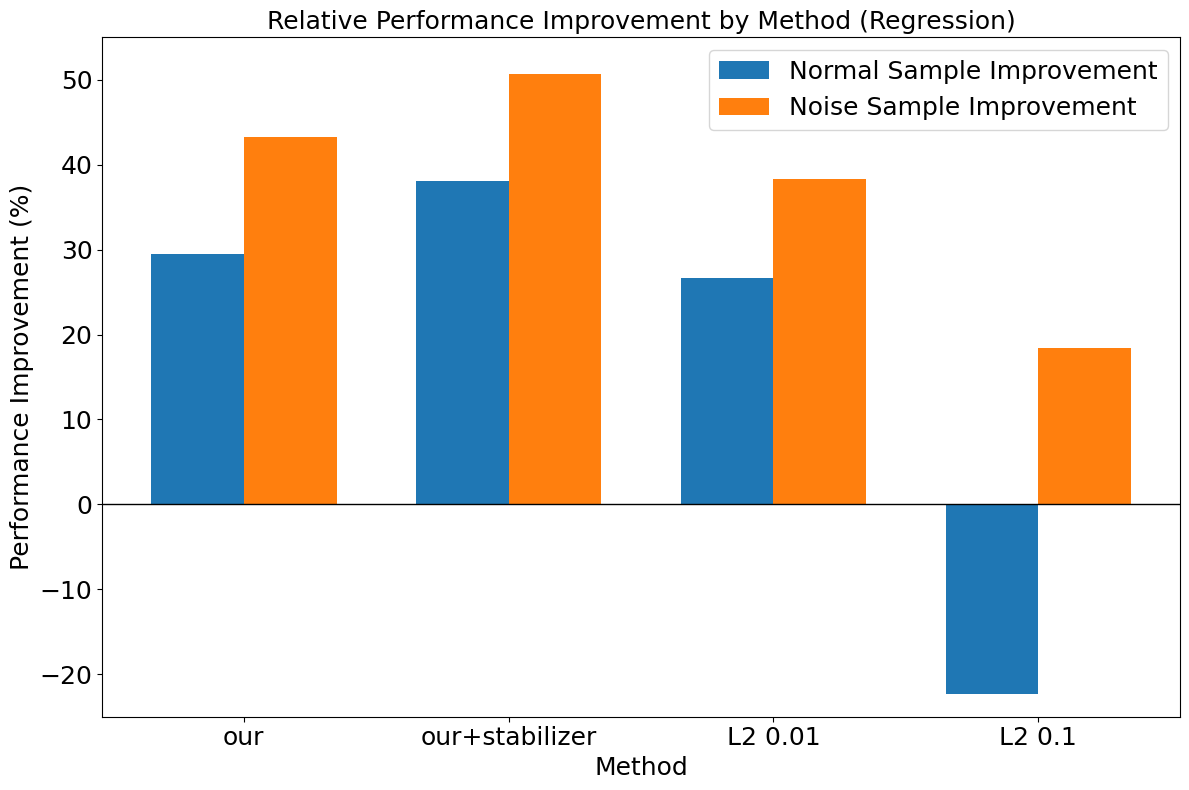}
    \includegraphics[width=0.75\linewidth]{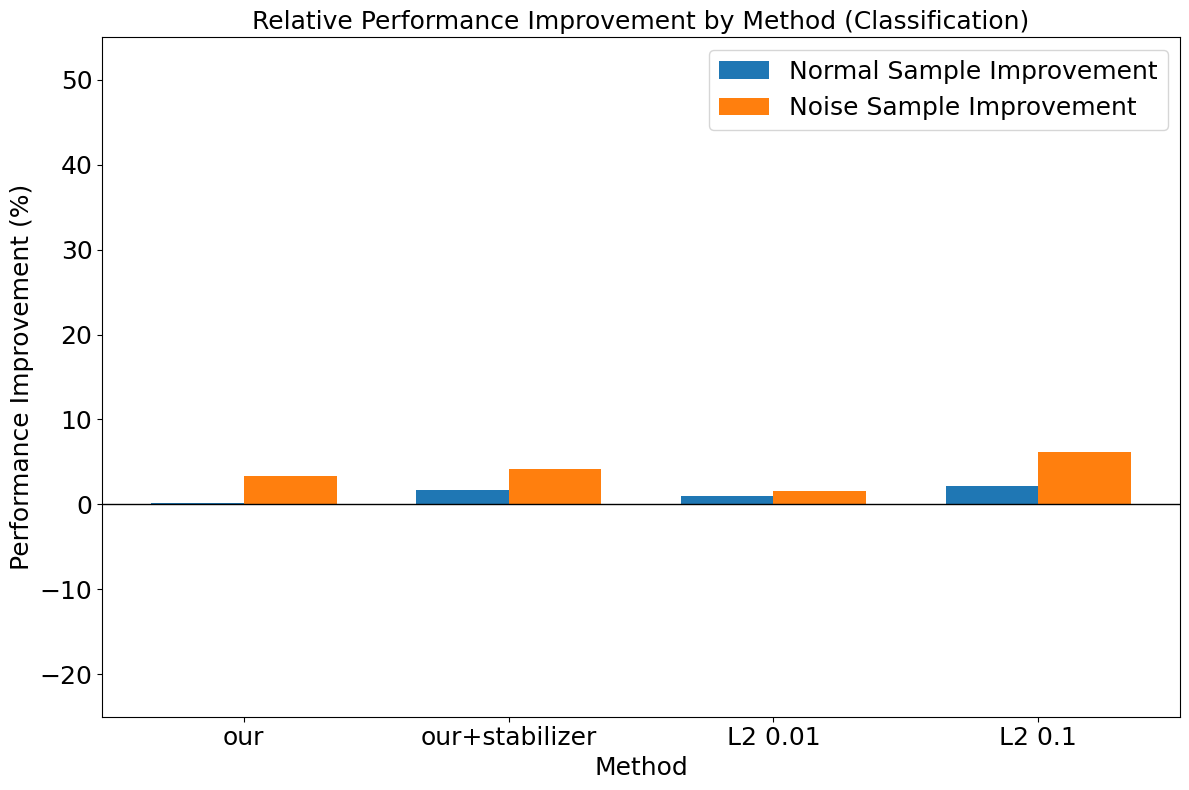}
    \caption{Relative performance improvement over the gradient method: (Top) regression task, (Bottom) classification task.}
    \label{fig:Regression_Classification}
\end{figure}



For the regression tasks, our method typically yielded lower loss than the baseline and is often better than L2 regularization. The L2 method’s effectiveness strongly depends on the regularization strength. For example, a high regularization factor (0.1) always forces parameters close to zero that the model outputs become trivial, resulting in poor performance uniformly across conditions. On the other hand, a weaker L2 strength (0.01) performs better in certain regression scenarios.

For the classification tasks, differences between methods are less dramatic. However, even here, the impact of regularization strength is evident. For instance, on some classification datasets, L2 with a strength of 0.1 outperforms L2 with 0.01, illustrating that the optimal regularization level is dataset-dependent.


As a final remark, we mention concerns about the stability of our method. While our approach generally surpasses the baseline in regression tasks and provides competitive performance in classification tasks, there are exceptions. For instance, on the CIFAR-10 dataset, we sometimes observe worse performance than the baseline (like figure \ref{fig:CIFAR10_results} with 3 layers and samples size 500). Further investigation revealed that this can occur when our method’s loss unexpectedly increases during later training epochs, leading to instability. We will address these instability issues in the subsequent discussion section, proposing potential solutions and adjustments to our method that could mitigate such occurrences.

\section{Discussion}

\begin{figure}[!]
    \centering
    \includegraphics[height=\textheight]{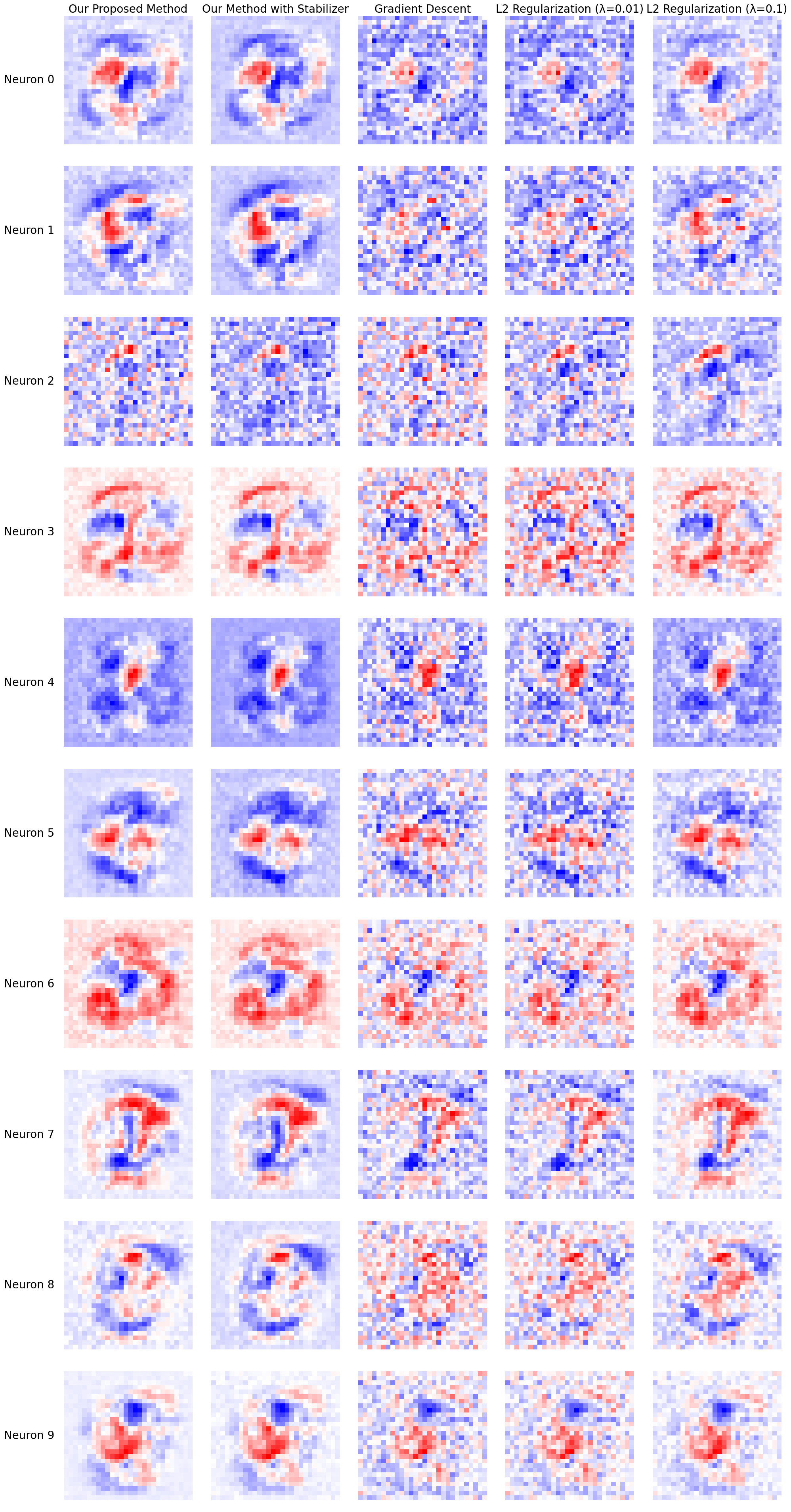}
    \caption{Visualization of receptive fields of neurons across models}
    \label{fig:receipt_field}
\end{figure}

\subsection{Experimental result discussion}
Overall, the proposed method demonstrated consistently decent performance without a need for tuning hyperparameters. While the baseline approach using plain backpropagation exhibited the highest error rates, L2 regularization sometimes outperformed both the baseline and the proposed method and sometimes underperformed the baseline. L2’s success was highly sensitive to the strength of the regularization parameter. Setting it too high led to underfitting—preventing the network from learning meaningful features—whereas too weak a setting reduced it to near-baseline results. By contrast, the proposed method did not depend as heavily on such fine-tuning. Although it did not always yield a strictly optimal solution, there was no evidence that it performed worse than the baseline, suggesting that it is at least a “harmless” choice.

We also observed task-dependent differences in their performance. The proposed method’s advantages were more pronounced in regression tasks with relatively low-dimensional input spaces (e.g., 8- or 11-dimensional data) with less redundancy, which makes noise strongly affect feature extraction. In these settings, restricting overfitting to noise had a meaningful positive impact on model performance. On the other hand, classification tasks involving image data showed less distinction among methods. Image inputs typically have higher dimensionality and exhibit spatial redundancy, which allows models to learn and leverage these redundancies to achieve robustness to noise. For instance, in a dataset like Fashion-MNIST, even under significant perturbations by independent noise, the underlying information encoded in the images remains distinguishable. Consequently, neither the proposed method nor L2 regularization provided a substantial advantage in this more noise-tolerant domain.

Although the numerical results of our method exhibited some similarity to those of the L2 regularization approach in the experimental outcomes, the two methods are fundamentally different. As illustrated in Figure \ref{fig:receipt_field}, we plotted the weight distributions of the receptive fields of neurons labeled 0 through 9 in the first layer of various models after training on the MNIST dataset with 1000 samples. Since all models started with the same initial weights, there is a degree of similarity in their resulting receptive fields despite differences in training methods. Notably, our method produces more distinct and noise-free receptive fields compared to gradient descent, which results in substantial noise in the receptive fields. While L2 regularization reduces noise to some extent as the regularization strength increases, it does not address the root cause of noise generation. This distinction underscores the fundamental differences between our method and L2 regularization. Although we conducted comparisons on relatively simple classification and regression tasks using small datasets, such tasks may not fully reveal the performance differences between our method and L2 regularization on more complex problems.

A key limitation of our approach is that it does not rely on enforcing strict gradient-based updates in each step. Instead, each neuron has a degree of autonomy in determining how it adjusts its weights, rather than adhering to the global gradient signal as strictly as in standard gradient descent. As a result, whether the model moves toward a more optimal solution can be somewhat random (like the result in the Cifar experiment). One way to mitigate this unpredictability is to employ batch training methods. By using multiple samples to guide the update direction, the model’s updates will become more stable and are more likely to move toward convergence. Of course, the exact design and implementation of such batch training approaches would be left to future work by those interested in refining this method.

Another drawback of the proposed method is that the neuron’s output range is effectively bounded by the range of the input $X$. Thus, large target values must be scaled down to fit into the neuron’s output range. Although this may initially seem like a disadvantage, it parallels biological constraints: both synaptic strengths and neuronal firing rates are inherently limited, preventing any single neuron from causing excessive damage if it “misfires.” From a practical standpoint, if a very large output is needed, rather than relying on a single neuron with disproportionately large weights, it is more robust to distribute the necessary capacity across multiple neurons. By aggregating their outputs, the network can produce the desired large value, and this architecture promotes greater fault tolerance. In addition to simply scaling down targets, one can also increase the number of neurons, effectively dispersing the load and enhancing the overall robustness of the model.

In summary, while our proposed method introduces certain complexities and constraints, it also offers new avenues for stable, noise-tolerant learning, especially in tasks sensitive to subtle feature variations. We remain confident that with further refinement—via batch training, optimized update rules, and careful architectural choices—this approach can become a highly competitive option for a wide range of learning scenarios.

\subsection{Comparison with contrastive learning (brief overview)}

While this work primarily focuses on decomposing each neuron’s weights into positive and negative features (\(W_1\) and \(W_2\)), the idea of learning by “contrast” is reminiscent of contrastive learning methods \cite{hadsell2006dimensionality,chen2020simclr,he2020moco}. In traditional contrastive learning, models aim to bring representations of similar data samples closer together and push dissimilar ones apart; this typically happens at the sample level—comparing input instances in pairs or batches. By contrast, our approach implements a neuron-level contrast that separates “what the neuron should detect” from “what it should ignore.” Although both strategies share the principle of leveraging contrasts, the scope and mechanism differ significantly. Readers interested in a more in-depth theoretical comparison, including how our neuron-centric perspective reduces reliance on large batch sizes and external contrast definitions, are directed to Appendix [A], where we explore these differences in greater detail.

\section{Code availability} The code used to analyze the data is available is available in the Github repository:
\\ \noindent
\href{https://github.com/Dataojitori/Dual-Weight-NN/tree/main}{https://github.com/Dataojitori/Dual-Weight-NN/tree/main}.

\section{Acknowledgments}

I would like to express my deep gratitude to ChatGPT for its invaluable assistance in refining the structure and articulation of this paper. It helped transform my initial, disorganized draft into a coherent and clearly articulated manuscript, making the ideas more accessible and easier to understand for readers. 

This work was supported by JSPS KAKENHI Grant Number JP 21H05246.

\bibliography{references} 

\newpage
\appendix

\section{Comparison with contrastive learning}

The idea that neural networks should learn from contrasts in the input is not unique to our research; it is a foundational principle in the field of contrastive learning. However, the key distinction lies in where and how this contrast is implemented. Contrastive learning~\cite{hadsell2006dimensionality,chen2020simclr} posits that the contrast should occur between different samples—specifically, positive and negative pairs—whereas our approach suggests that the contrast should occur within the neural network itself, particularly within the weights of each neuron.

\subsubsection{Overview of contrastive learning}

Contrastive learning is an unsupervised learning paradigm aimed at learning effective feature representations without relying on labeled data. Its core concept involves bringing similar samples closer together in the feature space while pushing dissimilar samples apart. This enables the neural network to automatically learn the intrinsic structure of the data.

Traditional methods like SimCLR~\cite{chen2020simclr} and MoCo~\cite{he2020moco} utilize data augmentation techniques to generate different views of the same data sample, treating these augmented versions as positive pairs. Different data samples serve as negative pairs. The neural network maps these samples into a latent space, and the loss function is designed to minimize the distance between positive pairs while maximizing the distance between negative pairs. This approach relies on predefined notions of which samples are positive and which are negative, and the learning process depends heavily on these assumptions.

Recent advancements like Bootstrap Your Own Latent~\cite{grill2020bootstrap} (BYOL) have introduced methods that do not require explicit negative pairs, overcoming some limitations of traditional contrastive learning. BYOL generates two augmented views of the same data sample and processes them through an online network and a target network. The objective is to minimize the difference between their representations to learn stable and useful feature representations. The target network's weights are updated using a momentum-based moving average of the online network's weights, rather than through direct gradient updates. While BYOL shares structural similarities with contrastive learning methods, its learning mechanism differs significantly from our proposed approach. Consequently, we have not included BYOL as a direct comparison in this paper.

\subsubsection{Key differences with traditional contrastive learning}

Our approach diverges from traditional contrastive learning in several important aspects:

\textbf{Contrast at the neuron level} In traditional contrastive learning, contrast occurs globally at the sample level through the loss function, which considers relationships between different samples or sample pairs. In our method, contrast occurs locally within each neuron. Each neuron maintains separate positive (\( W_1 \)) and negative (\( W_2 \)) weight components and updates them based on its individual input and feedback. This means that within the same training iteration, some neurons may interpret the current input as positive while others may interpret it as negative, all based on their internal states and criteria.

This neuron-level autonomy allows each neuron to independently decide whether to treat the current input as a positive or negative example. The decision is based on internal factors, such as the sign of its gradient, rather than being dictated by external sample labels. This contrasts with traditional contrastive learning, where the designation of positive and negative samples is determined before training and uniformly applied across the network. Our method enables more fine-grained control over the learning process, allowing for a finer granularity of adaptation and potentially capturing more nuanced patterns in the data.

\textbf{Reduced dependence on large batch sizes} A significant challenge in traditional contrastive learning is the reliance on large batch sizes or memory banks to provide sufficient negative samples for effective learning. The performance of these methods often improves with the number of negative samples used in each update because the contrastive loss depends on comparing multiple samples simultaneously. In contrast, our method does not rely on contrasts between different samples within a batch. Instead, the contrast is internal to each neuron, occurring between its positive and negative weight components accumulated over time. As a result, our approach can perform effectively even with small batch sizes, including a batch size of one, as demonstrated in our experiments. This reduces computational overhead and memory requirements, making our method more practical in scenarios with limited resources.

\textbf{Flexibility in supervised and unsupervised settings} While we utilize supervised learning in our experiments for clarity and ease of demonstration, our approach is not limited to supervised settings. Fundamentally, our method divides the feedback received by each neuron's output into two states—positive and negative—and derives the neuron's weights from the difference between the average input vectors in these two states. 

In supervised learning, we define the boundary between these two states based on the sign of the neuron's gradient, which is informed by the labeled data. In unsupervised learning, we can define this boundary using alternative criteria. For example, if we aim to maintain a consistent total activation within a layer, we can set a threshold for the layer's total output. When the total output exceeds this threshold, neurons may treat the current input as negative to reduce activation; when it is below the threshold, they may treat it as positive to increase activation. This adaptability allows our method to be applied in various learning scenarios without being constrained by the availability of labeled data.


In summary, traditional contrastive learning relies on predefined notions of similarity and dissimilarity between samples, using global sample-level contrasts to guide learning. The designation of positive and negative pairs is determined externally and remains fixed during training. In contrast, our method shifts the focus to neuron-level contrasts, enabling each neuron to autonomously determine and learn from the differences between inputs it considers positive or negative. This internal contrast mechanism reduces the dependence on external labels and large batches of samples, offering a more granular and flexible approach to learning.

By empowering neurons to independently contrast their inputs and update their weights accordingly, our method provides a novel perspective on how neural networks can learn from data. This contrasts with traditional contrastive learning, where contrasts are defined globally and applied uniformly across the network. Our approach suggests that introducing contrast within the neural network's internal structure can lead to more efficient and adaptable learning mechanisms, potentially enhancing the network's ability to generalize from limited or unbalanced data.

\section{Proposed method augmented by stabilizer}

First, we calculate the updaterate based on the gradient value, which determines how strongly we adjust the weights. 
\[
\text{updaterate} = \frac{|\text{grad}_i|}{G_{\text{avg}}} \times 0.1, \quad \text{capped at } 1
\]
Then, we use this updaterate to update $W_1$ and $W_2$ accordingly, depending on the gradient's sign. 
\[
\begin{cases}
    \text{If } \text{grad}_i < 0:& W_1^{\text{new}} = W_1^{\text{old}} \times (1 - \eta \times \text{updaterate}) + X_i \times \eta \times \text{updaterate} \\\\
    \text{If } \text{grad}_i > 0:& W_2^{\text{new}} = W_2^{\text{old}} \times (1 - \eta \times \text{updaterate}) + X_i \times \eta \times \text{updaterate} \\\\
    \text{If } \text{grad}_i = 0:& W_1^{\text{new}} = W_1^{\text{old}}, \quad W_2^{\text{new}} = W_2^{\text{old}}
\end{cases}
\]
Finally, we update $G_{\text{avg}}$ to maintain a moving average of the gradient, ensuring a smoother representation: 
\[
\text{If } \text{grad}_i \neq 0: \quad G_{\text{avg}}^{\text{new}} = G_{\text{avg}}^{\text{old}} \times (1 - \eta) + |\text{grad}_i| \times \eta.
\]
Here we used the same learning rate used in the weight update.

\section{Proof that the expected value of the moving average \( W_n \) converges to \( E[X] \):}
\label{sec:convergence_proof}

In this section, we provide a step-by-step proof to demonstrate that the expected value of the moving average \( W_n \) converges to \( E[X] \), the expected value of \( X \). To facilitate understanding, we include explanations at each step.

1. \textbf{Moving Average Update Equation:}

The moving average update equation is given by:
\begin{align}
W_{n+1} = W_n (1 - a_n) + X_n a_n,
\end{align}
where:
\begin{itemize}
    \item \( W_n \) is the weight vector after \( n \) updates.
    \item \( X_n \) is a random variable with \( E[X_n] = E[X] \).
    \item \( a_n = \eta |\text{grad}_n| \), \( 0 < a_n < 1 \), \( \eta > 0 \).
    \item \( \text{grad}_n < 0 \), \( \lim_{n \to \infty} \text{grad}_n = 0 \).
\end{itemize}

2. \textbf{Deviation from the Expected Value:}

To analyze convergence, we focus on the deviation of \( W_n \) from \( E[X] \). Subtract \( E[X] \) from both sides of the update equation:

\begin{align}
W_{n+1} - E[X] = (1 - a_n)(W_n - E[X]) + a_n (X_n - E[X]).
\end{align}

3. \textbf{Expectation of the Deviation:}

Taking the expectation of both sides, we get:

\begin{align}
E[W_{n+1} - E[X]] = (1 - a_n) E[W_n - E[X]] + a_n E[X_n - E[X]].
\end{align}

Since \( E[X_n - E[X]] = 0 \), this simplifies to:

\begin{align}
E[W_{n+1} - E[X]] = (1 - a_n) E[W_n - E[X]].
\end{align}

4. \textbf{Recursive Relation:}

By iterating the above relation over multiple steps, we obtain:

\begin{align}
E[W_{n+1} - E[X]] = \left( \prod_{k=1}^n (1 - a_k) \right) (W_0 - E[X]).
\end{align}

5. \textbf{Convergence Condition:}

To ensure that \( E[W_{n+1} - E[X]] \to 0 \) as \( n \to \infty \), we require:

\begin{align}
\sum_{k=1}^\infty a_k = \infty.
\end{align}

This condition implies that the step sizes \( a_k \) decrease slowly enough such that their cumulative sum diverges, allowing sufficient updates to eliminate the deviation over time. Consequently:

\begin{align}
\prod_{k=1}^\infty (1 - a_k) = 0.
\end{align}

6. \textbf{Conclusion:}

Under the condition \( \sum_{k=1}^\infty a_k = \infty \), we conclude that:

\begin{align}
\lim_{n \to \infty} E[W_{n+1}] = E[X].
\end{align}

This demonstrates that the moving average \( W_n \) converges in expectation to the true expected value \( E[X] \), provided the step sizes \( a_k \) satisfy the specified conditions.

\section{Proof that the expected value of the weighted average \( W_1 \) is \( \eta E[X] \):}
\label{sec:convergence_proof2}

Here, we provide a concise proof to show that the expected value of the weighted average \( W_1 \) converges to \( \eta E[X] \).

1. \textbf{Weighted Average Formula:}

The weighted average is defined as:

\begin{align}
W_1 = \frac{\eta}{\sum_{\text{grad}_i < 0} (-\text{grad}_i)} \sum_{\text{grad}_i < 0} (-\text{grad}_i) X_i.
\end{align}

This formula uses \( \eta \) as a scaling factor and weights each \( X_i \) by the corresponding negative gradient magnitude \( -\text{grad}_i \). 

2. \textbf{Simplifying the Expression:}

Define \( S = \sum_{\text{grad}_i < 0} (-\text{grad}_i) \). Substituting \( S \) into the equation:

\begin{align}
W_1 = \eta \left( \frac{\sum_{\text{grad}_i < 0} (-\text{grad}_i) X_i}{S} \right).
\end{align}

3. \textbf{Express \( X_i \) in Terms of \( E[X] \):}

We decompose \( X_i \) into its expected value and a zero-mean error term:
\begin{align}
X_i = E[X] + \epsilon_i,
\end{align}
where \( \epsilon_i \) is the deviation of \( X_i \) from its expected value, and \( E[\epsilon_i] = 0 \).

4. \textbf{Substitute into \( W_1 \):}

Substituting \( X_i = E[X] + \epsilon_i \) into the expression for \( W_1 \), we get:

\begin{align}
W_1 = \eta \left( E[X] + \frac{\sum_{\text{grad}_i < 0} (-\text{grad}_i) \epsilon_i}{S} \right).
\end{align}

5. \textbf{Analyze the Second Term:}

The second term is the weighted average of \( \epsilon_i \):

\begin{align}
\frac{\sum_{\text{grad}_i < 0} (-\text{grad}_i) \epsilon_i}{S}.
\end{align}

Since \( E[\epsilon_i] = 0 \), we have:

\begin{align}
E\left[ \frac{\sum_{\text{grad}_i < 0} (-\text{grad}_i) \epsilon_i}{S} \right] = 0.
\end{align}

6. \textbf{Conclusion:}

Assuming the second term is negligible as \( i \to \infty \), we conclude:

\begin{align}
E[W_1] = \eta E[X].
\end{align}

This concludes the proof, showing that the expected value of \( W_1 \) is proportional to \( E[X] \) with a factor \( \eta \).



\section{Training plot}

\newgeometry{landscape, left=1in, right=1in, top=1in, bottom=1in}

\clearpage

\begin{landscape}
\begin{figure}[h!]
\begin{minipage}{0.33\linewidth}
\centering
\includegraphics[width=0.95\linewidth]{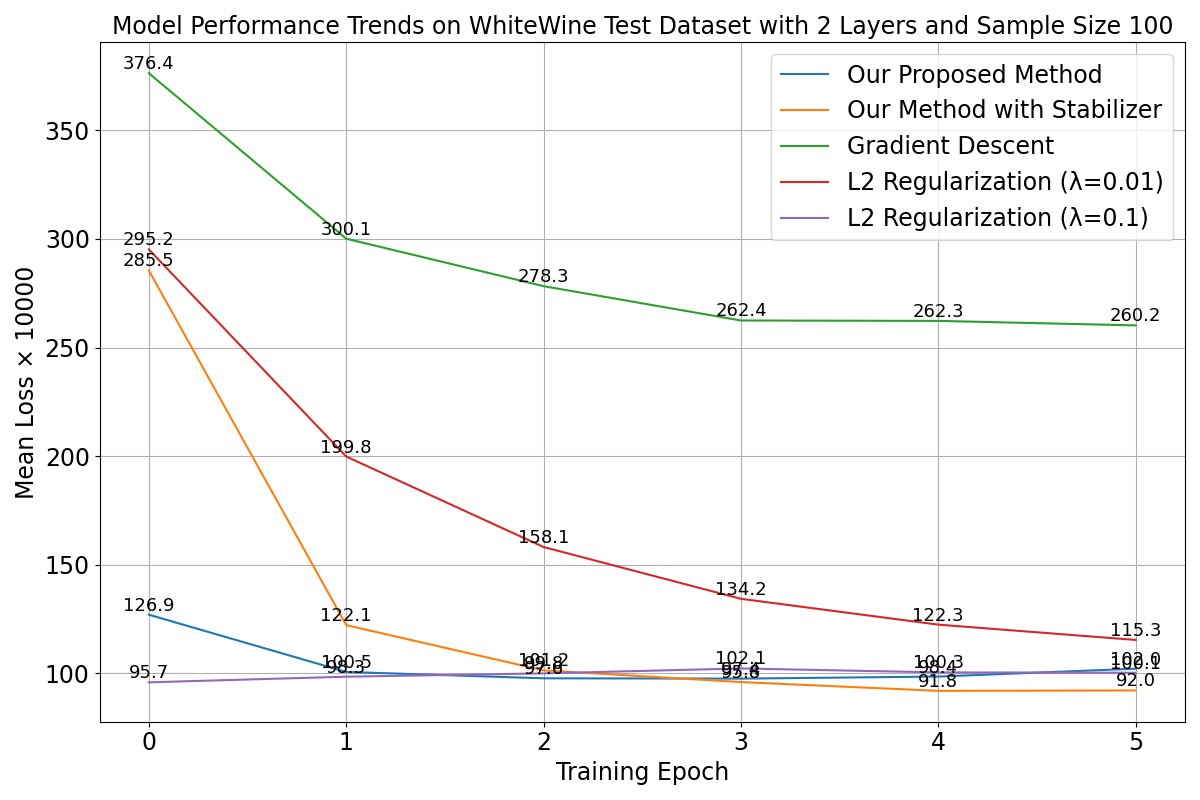}
\label{fig:WhiteWine_layers2_samples100}
\end{minipage}
\begin{minipage}{0.33\linewidth}
\centering
\includegraphics[width=0.95\linewidth]{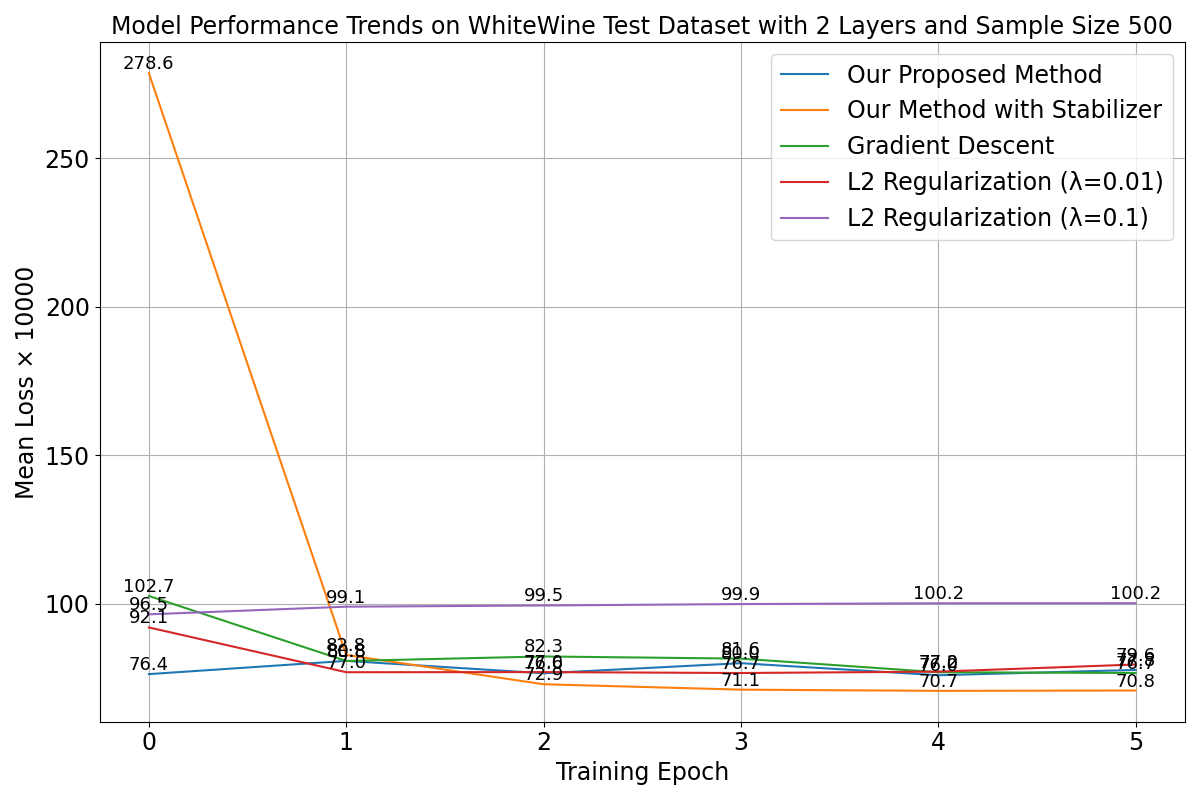}
\label{fig:WhiteWine_layers2_samples500}
\end{minipage}
\begin{minipage}{0.33\linewidth}
\centering
\includegraphics[width=0.95\linewidth]{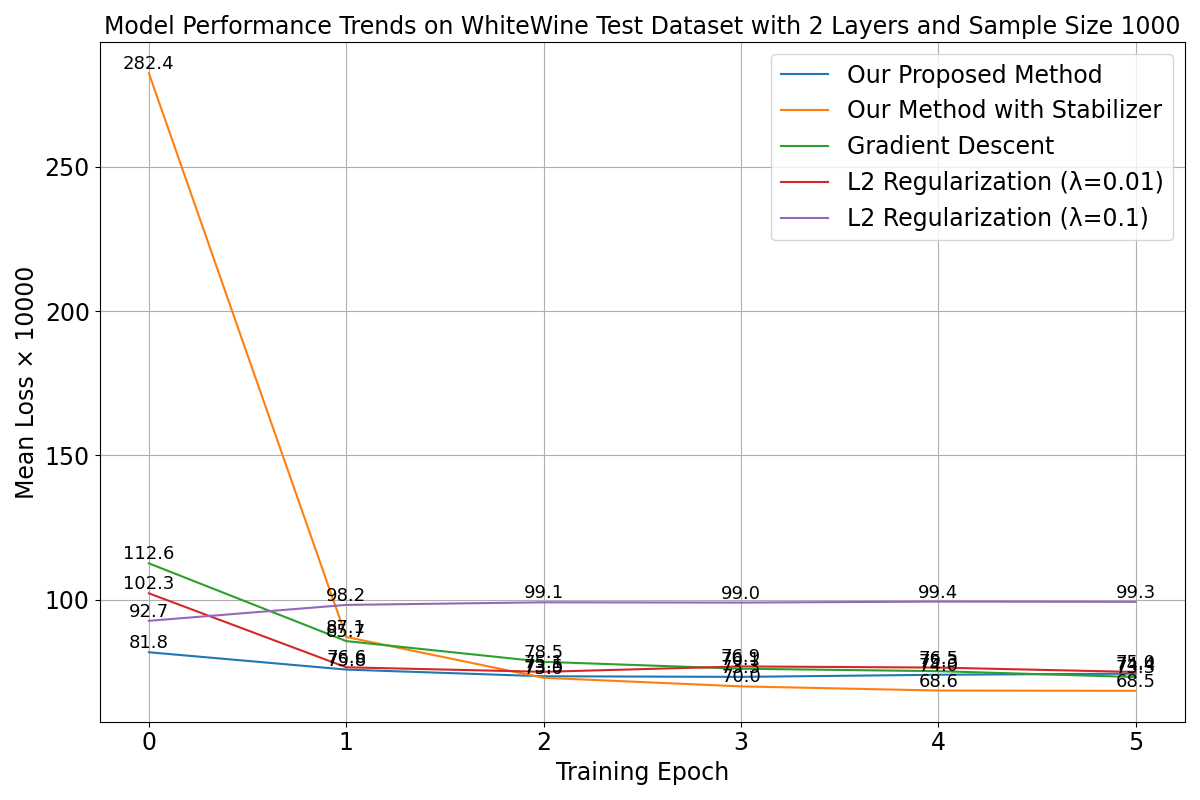}
\label{fig:WhiteWine_layers2_samples1000}
\end{minipage}
\\\\
\begin{minipage}{0.33\linewidth}
\centering
\includegraphics[width=0.95\linewidth]{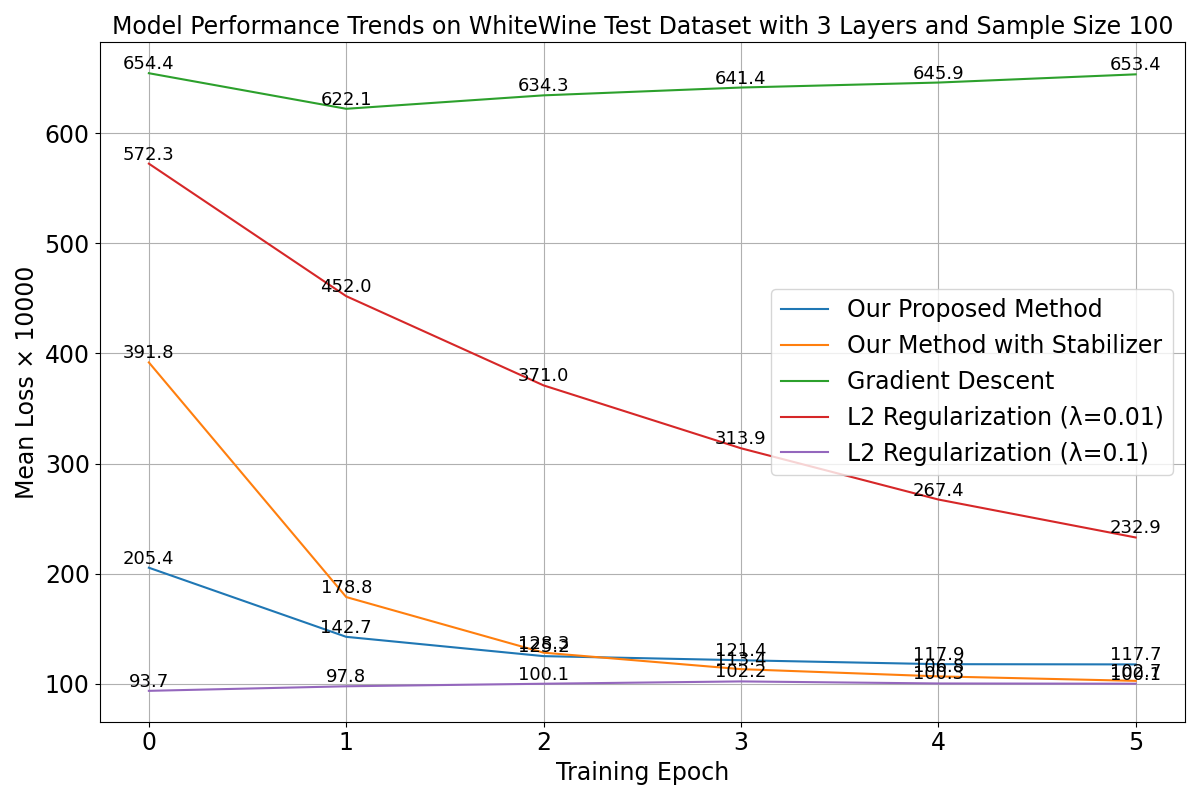}
\label{fig:WhiteWine_layers3_samples100}
\end{minipage}
\begin{minipage}{0.33\linewidth}
\centering
\includegraphics[width=0.95\linewidth]{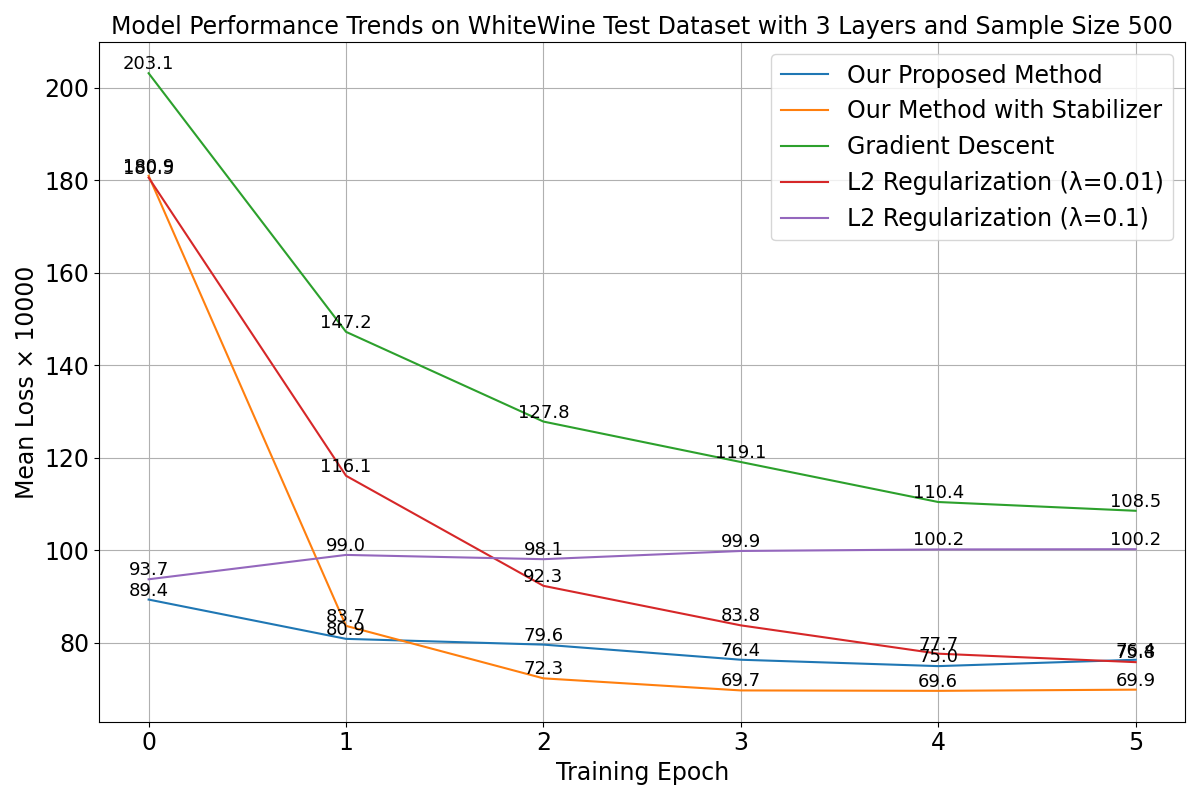}
\label{fig:WhiteWine_layers3_samples500}
\end{minipage}
\begin{minipage}{0.33\linewidth}
\centering
\includegraphics[width=0.95\linewidth]{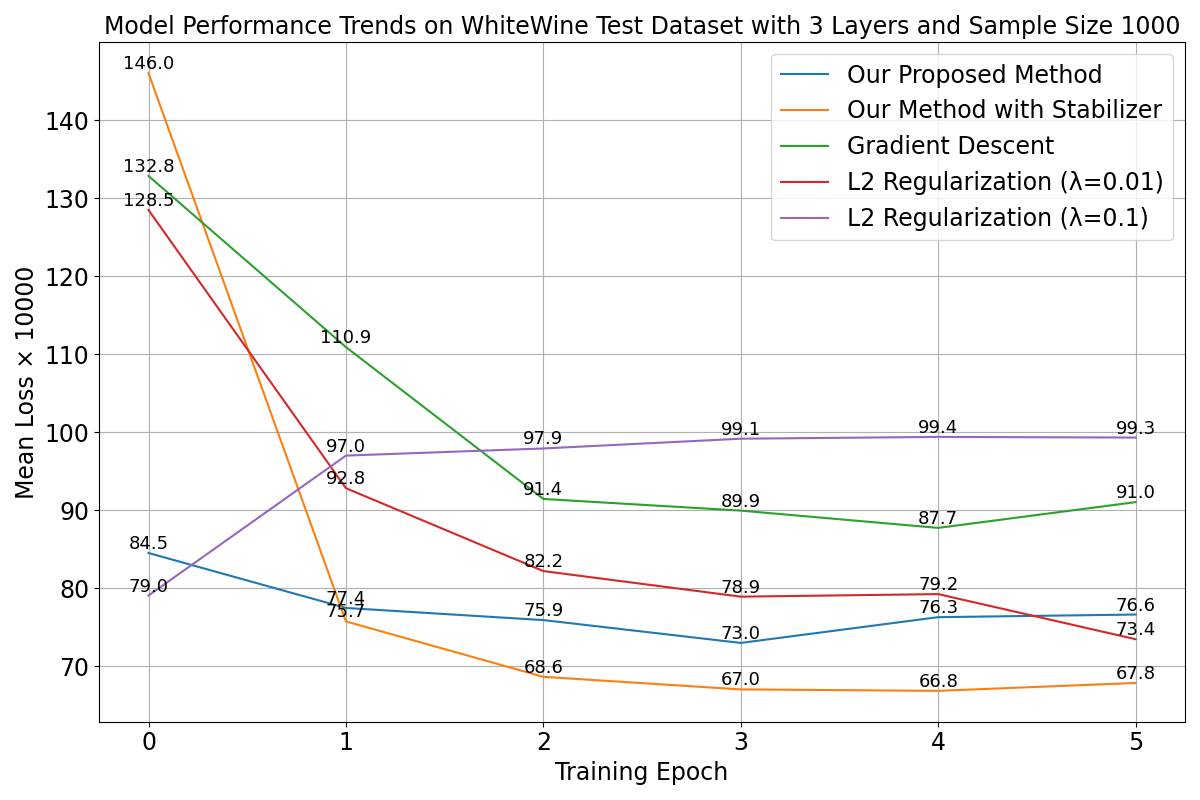}
\label{fig:WhiteWine_layers3_samples1000}
\end{minipage}
\caption{Results on WhiteWine dataset with models with 2 layers (Top row) and 3 layers (Bottom row). From left to right: Sample Size 100, 500, 1000.}
\end{figure}

\begin{figure}[h!]
\begin{minipage}{0.33\linewidth}
\centering
\includegraphics[width=0.95\linewidth]{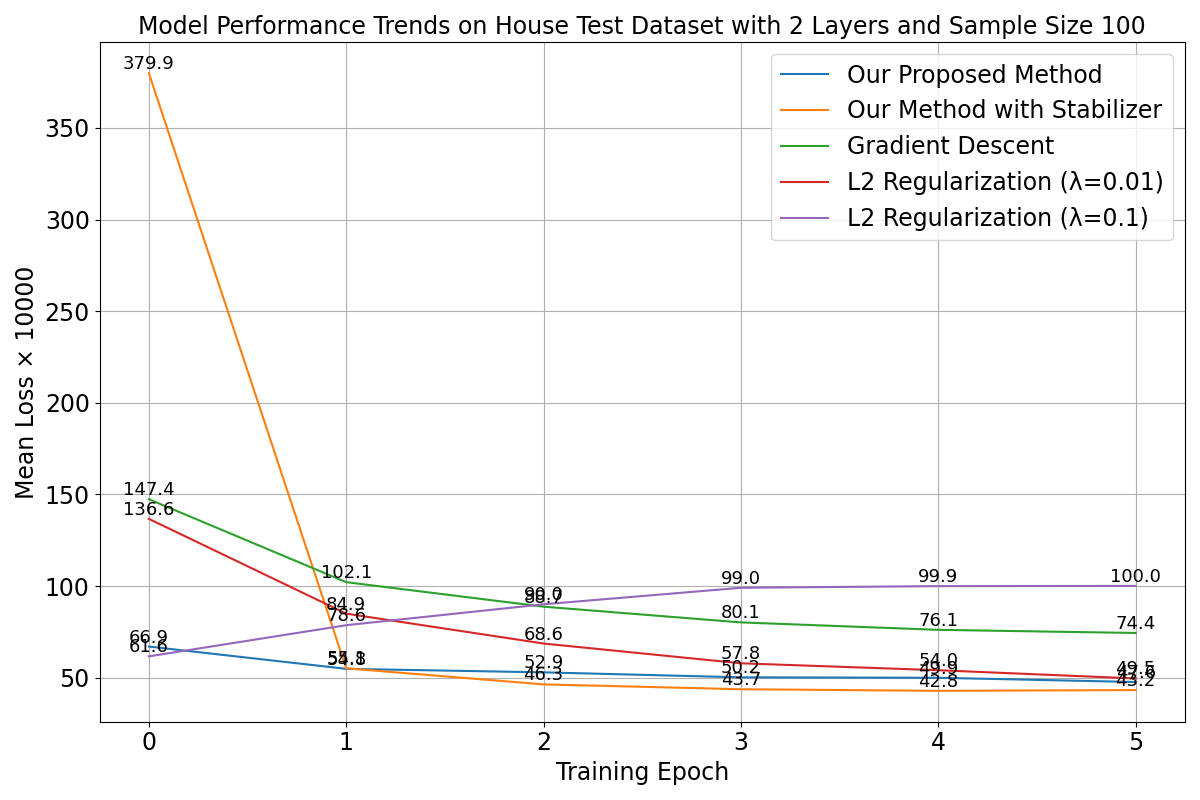}
\label{fig:House_layers2_samples100}
\end{minipage}
\begin{minipage}{0.33\linewidth}
\centering
\includegraphics[width=0.95\linewidth]{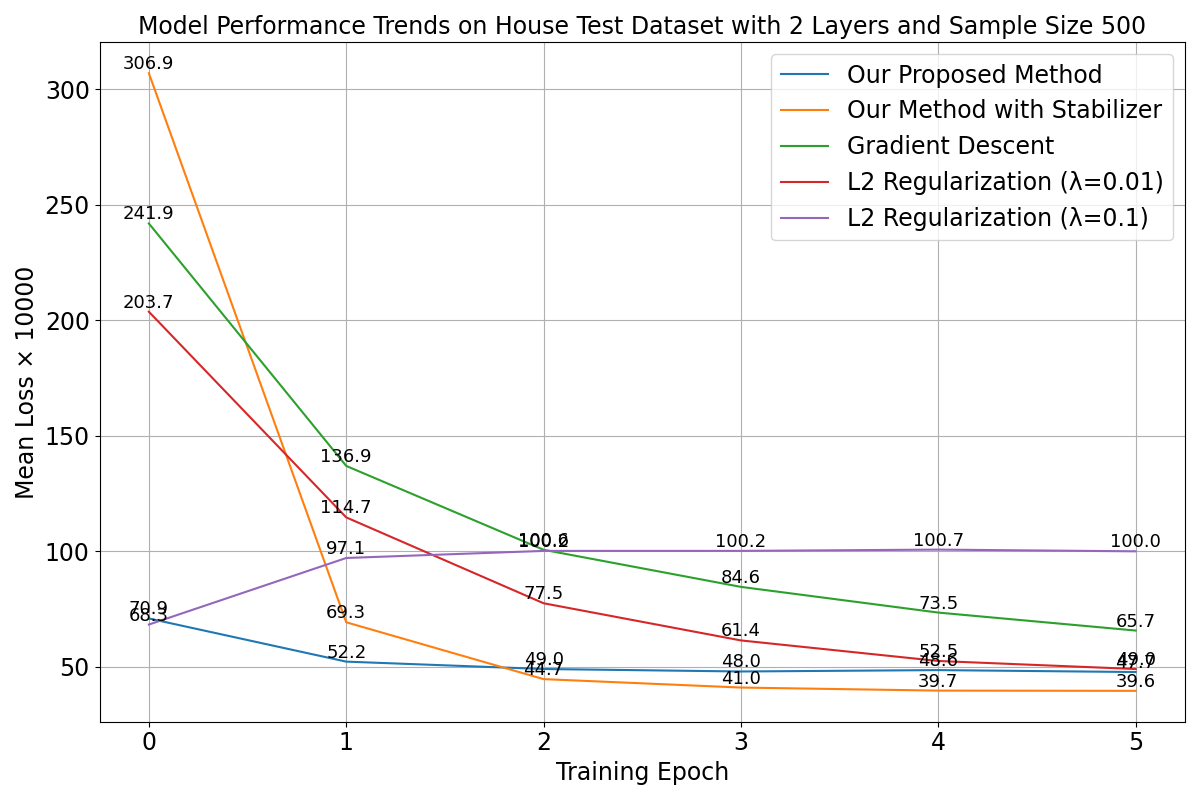}
\label{fig:House_layers2_samples500}
\end{minipage}
\begin{minipage}{0.33\linewidth}
\centering
\includegraphics[width=0.95\linewidth]{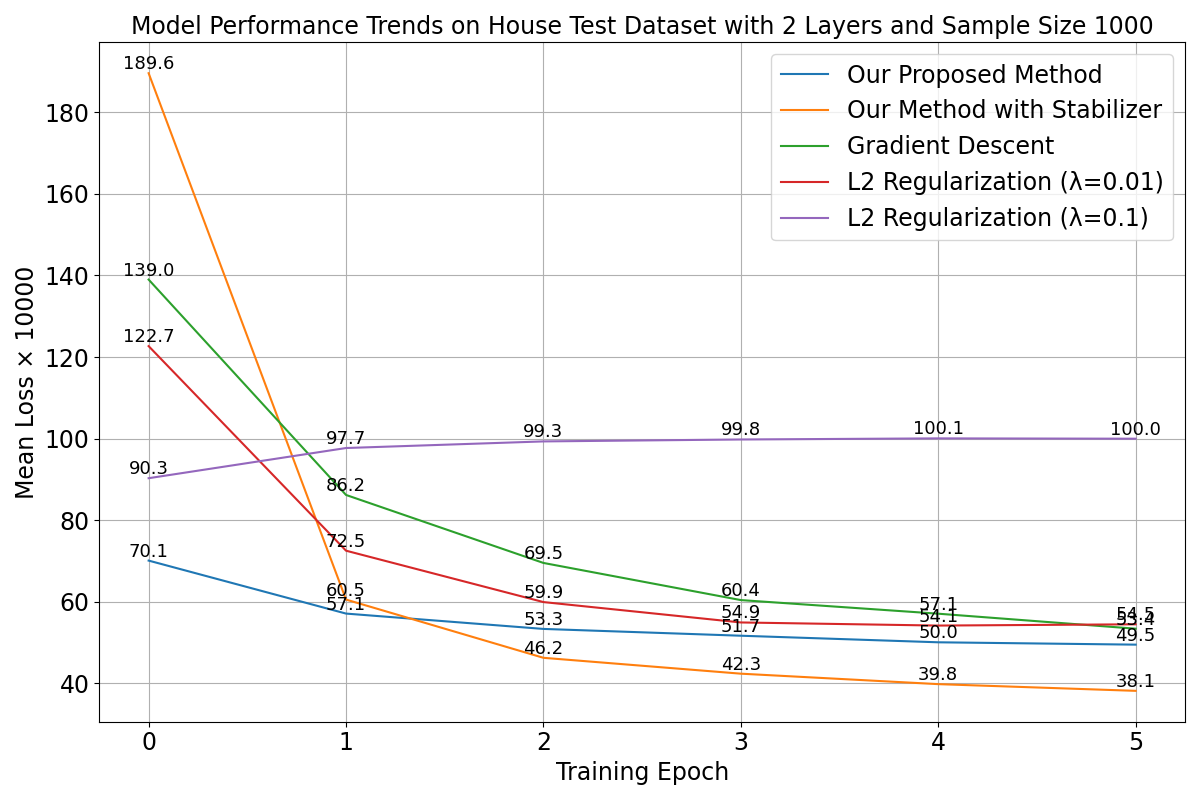}
\label{fig:House_layers2_samples1000}
\end{minipage}
\\
\begin{minipage}{0.33\linewidth}
\centering
\includegraphics[width=0.95\linewidth]{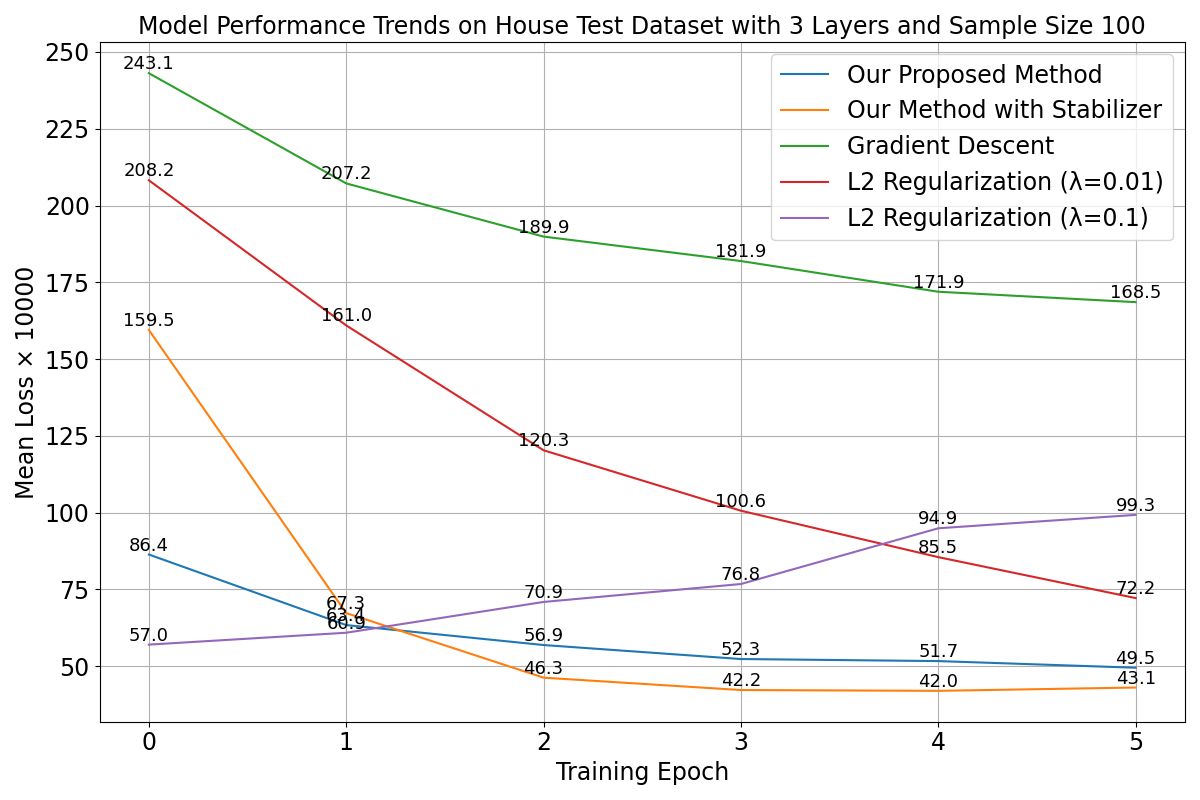}
\label{fig:House_layers3_samples100}
\end{minipage}
\begin{minipage}{0.33\linewidth}
\centering
\includegraphics[width=0.95\linewidth]{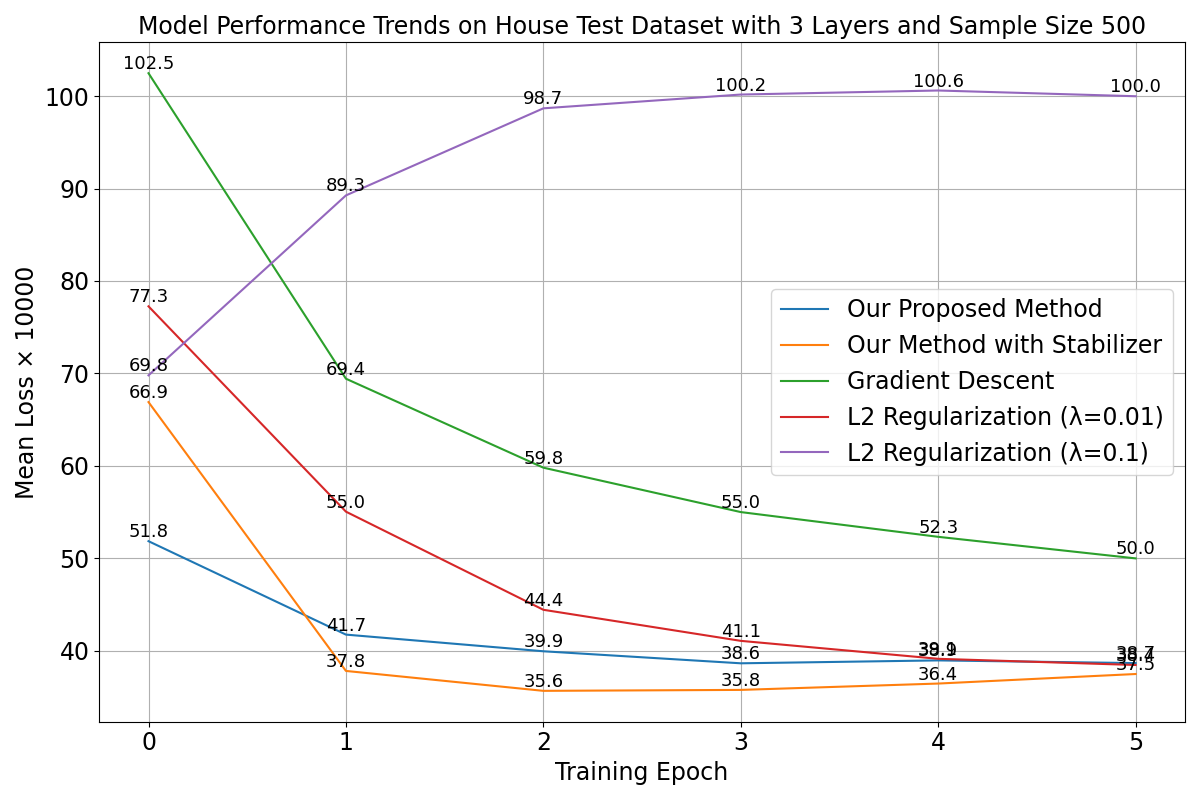}
\label{fig:House_layers3_samples500}
\end{minipage}
\begin{minipage}{0.33\linewidth}
\centering
\includegraphics[width=0.95\linewidth]{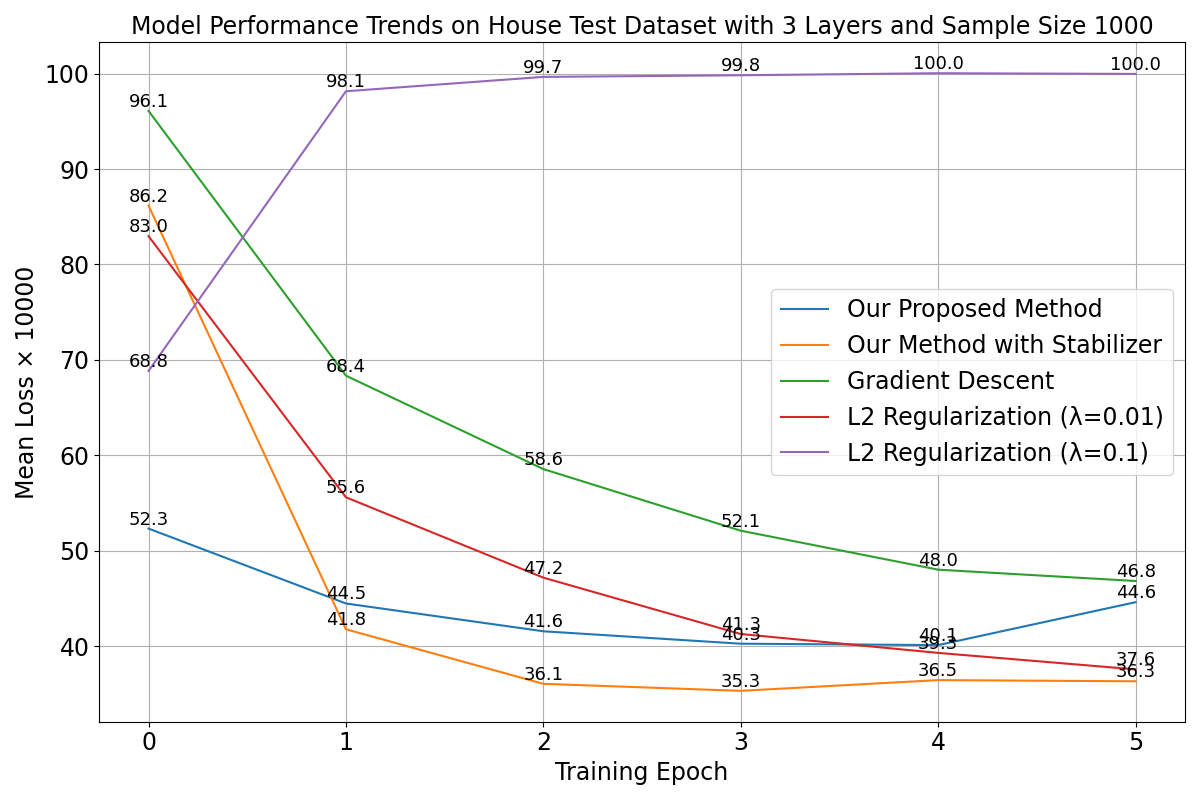}
\label{fig:House_layers3_samples1000}
\end{minipage}
\caption{Results on House dataset with models with 2 layers (Top row) and 3 layers (Bottom row). From left to right: Sample Size 100, 500, 1000.}
\end{figure}

\begin{figure}[h!]
\begin{minipage}{0.33\linewidth}
\centering
\includegraphics[width=0.95\linewidth]{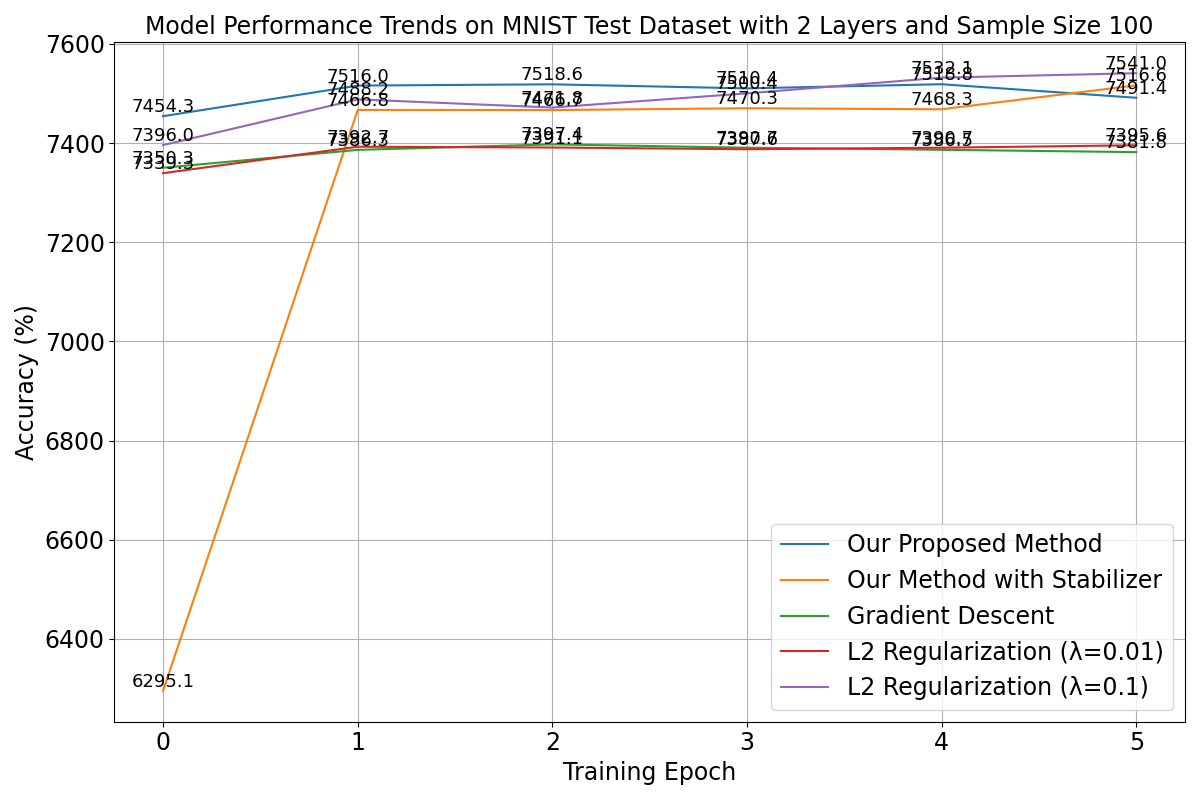}
\label{fig:MNIST_layers2_samples100}
\end{minipage}
\begin{minipage}{0.33\linewidth}
\centering
\includegraphics[width=0.95\linewidth]{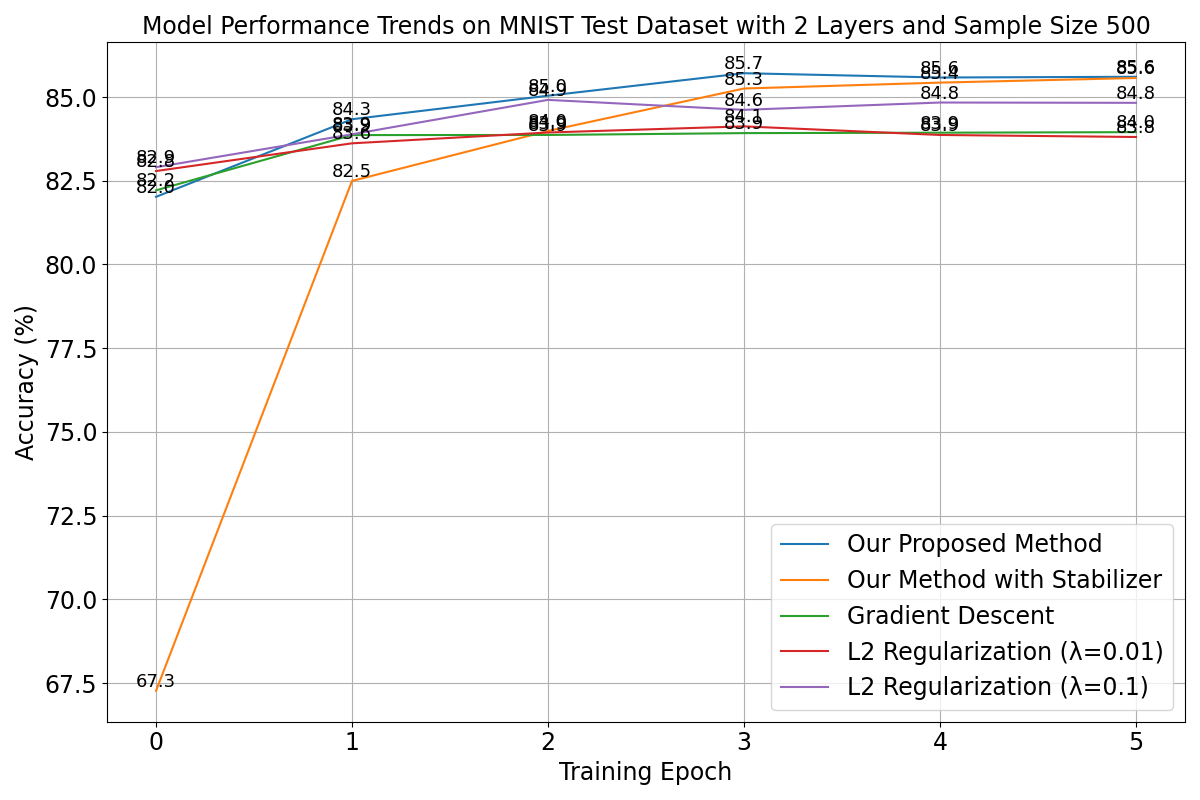}
\label{fig:MNIST_layers2_samples500}
\end{minipage}
\begin{minipage}{0.33\linewidth}
\centering
\includegraphics[width=0.95\linewidth]{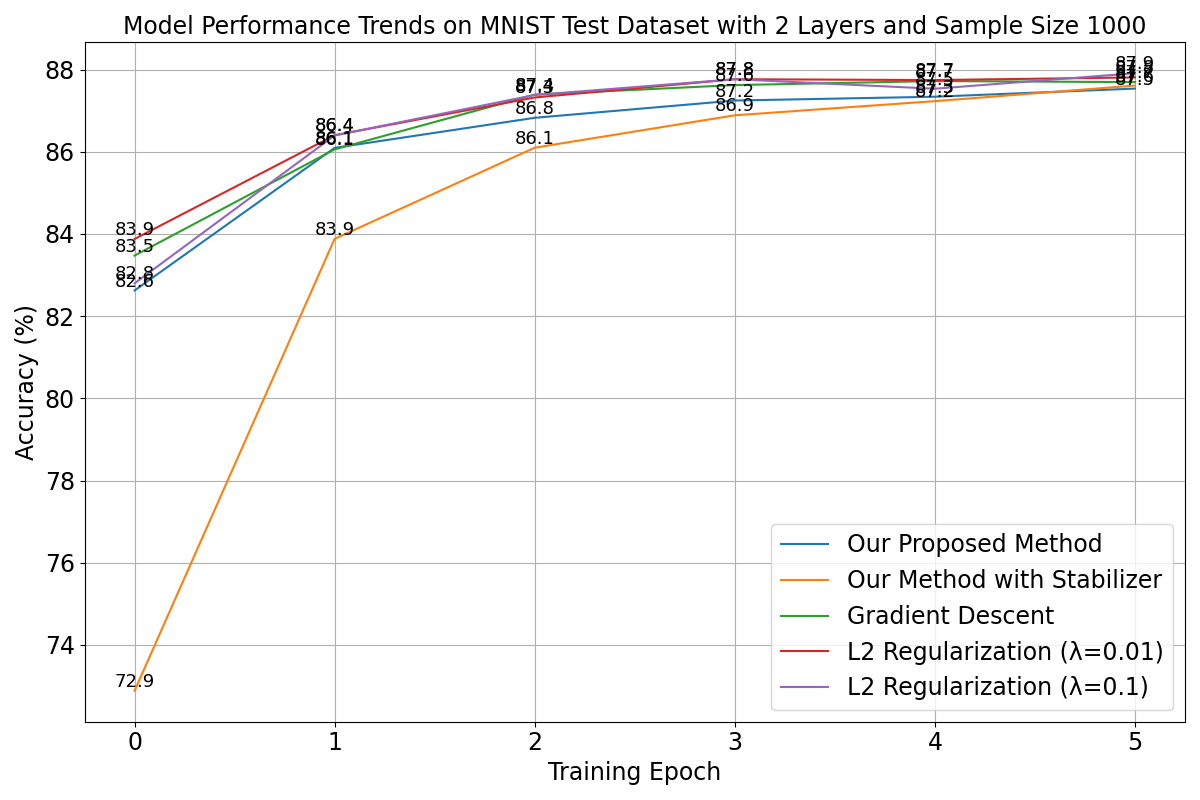}
\label{fig:MNIST_layers2_samples1000}
\end{minipage}
\\
\begin{minipage}{0.33\linewidth}
\centering
\includegraphics[width=0.95\linewidth]{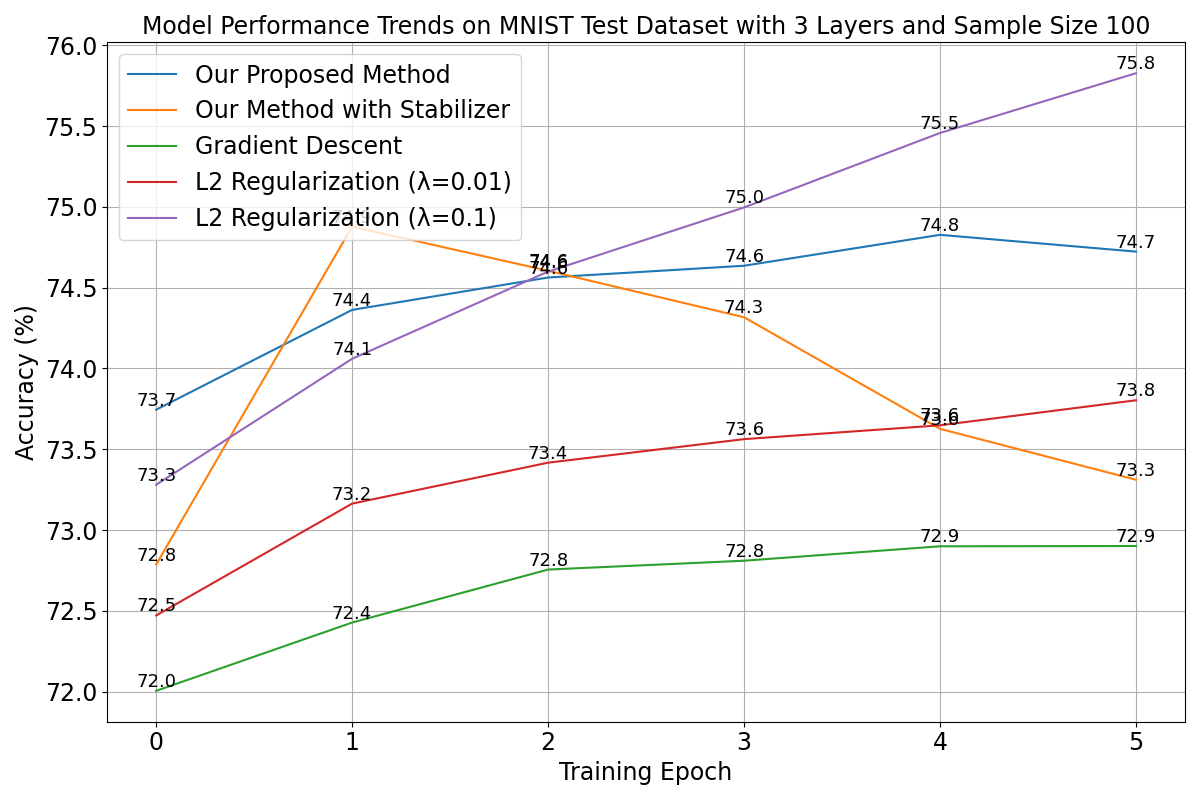}
\label{fig:MNIST_layers3_samples100}
\end{minipage}
\begin{minipage}{0.33\linewidth}
\centering
\includegraphics[width=0.95\linewidth]{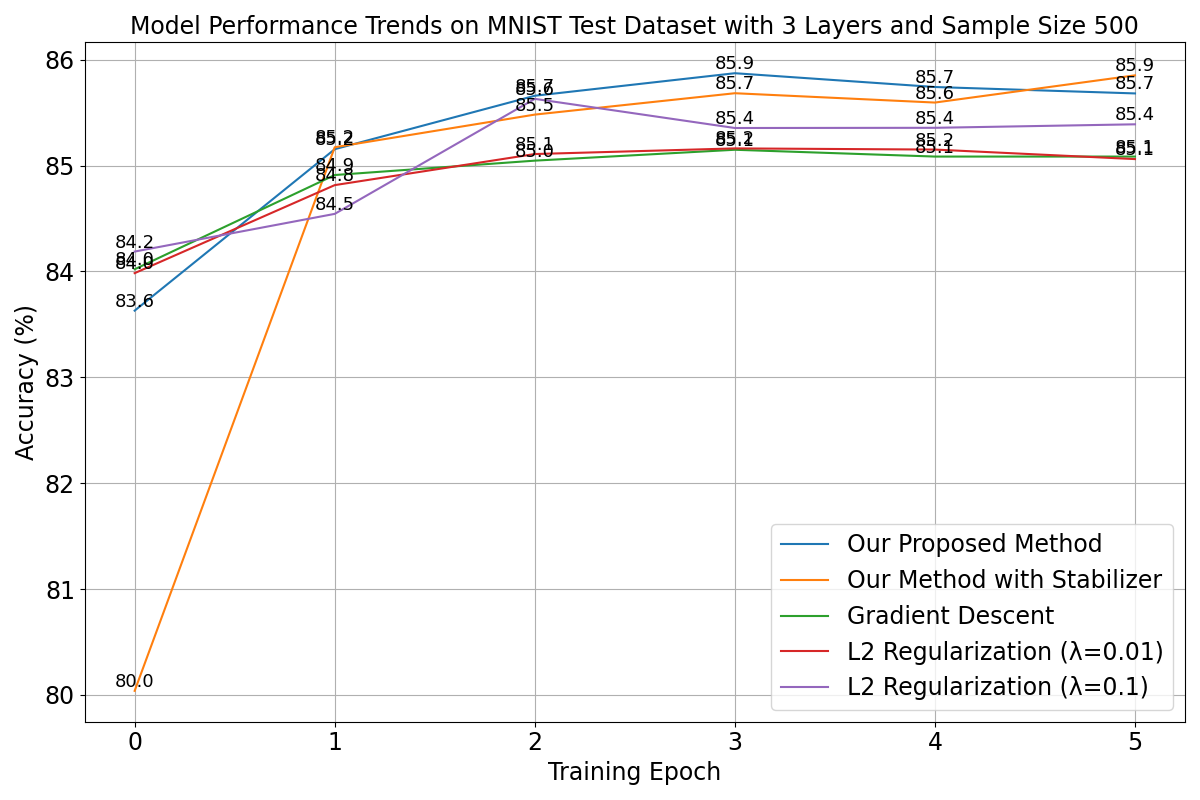}
\label{fig:MNIST_layers3_samples500}
\end{minipage}
\begin{minipage}{0.33\linewidth}
\centering
\includegraphics[width=0.95\linewidth]{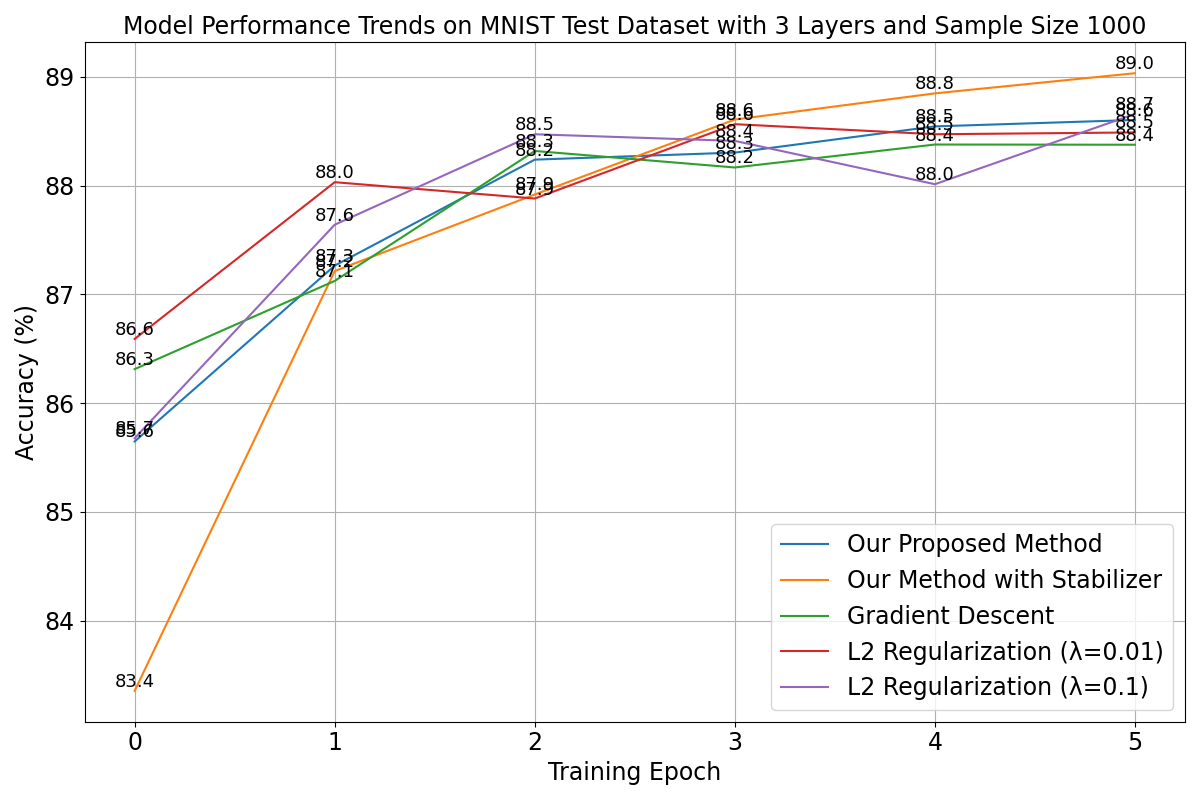}
\label{fig:MNIST_layers3_samples1000}
\end{minipage}
\caption{Results on MNIST dataset with models with 2 layers (Top row) and 3 layers (Bottom row). From left to right: Sample Size 100, 500, 1000.}
\end{figure}

\begin{figure}[h!]
\begin{minipage}{0.33\linewidth}
\centering
\includegraphics[width=0.95\linewidth]{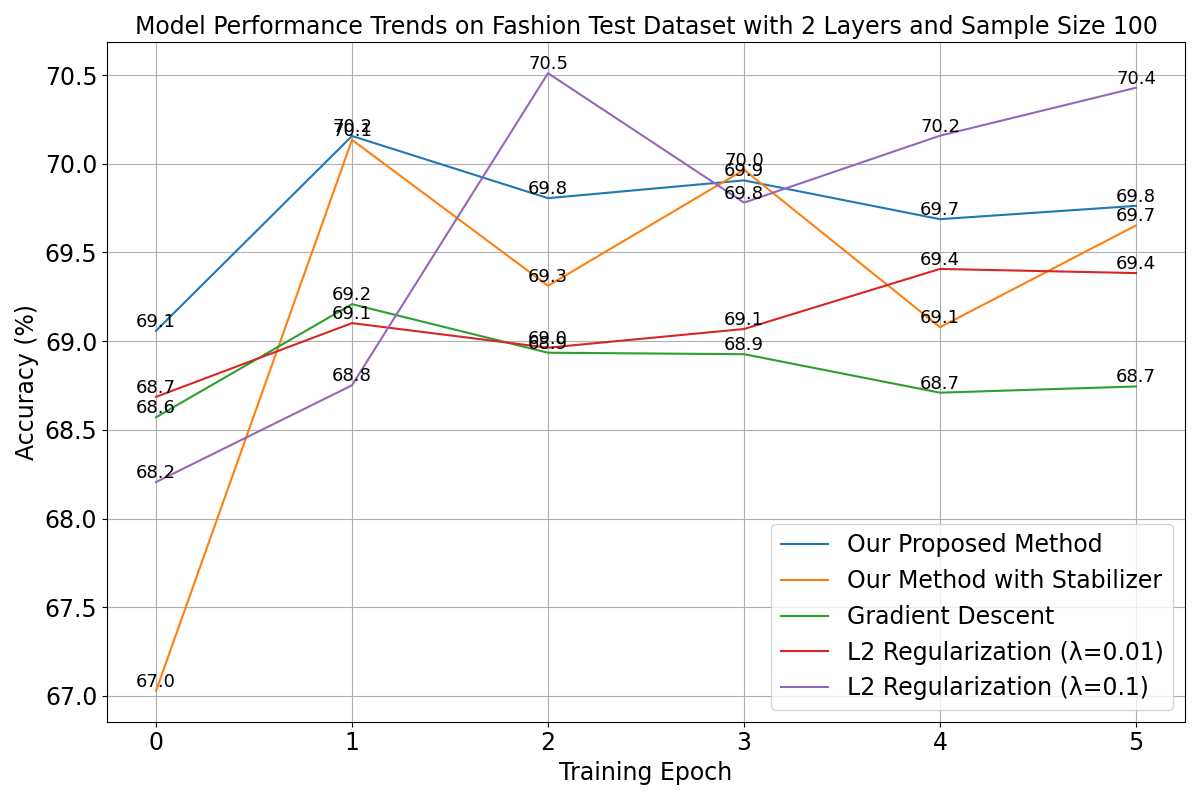}
\label{fig:Fashion_layers2_samples100}
\end{minipage}
\begin{minipage}{0.33\linewidth}
\centering
\includegraphics[width=0.95\linewidth]{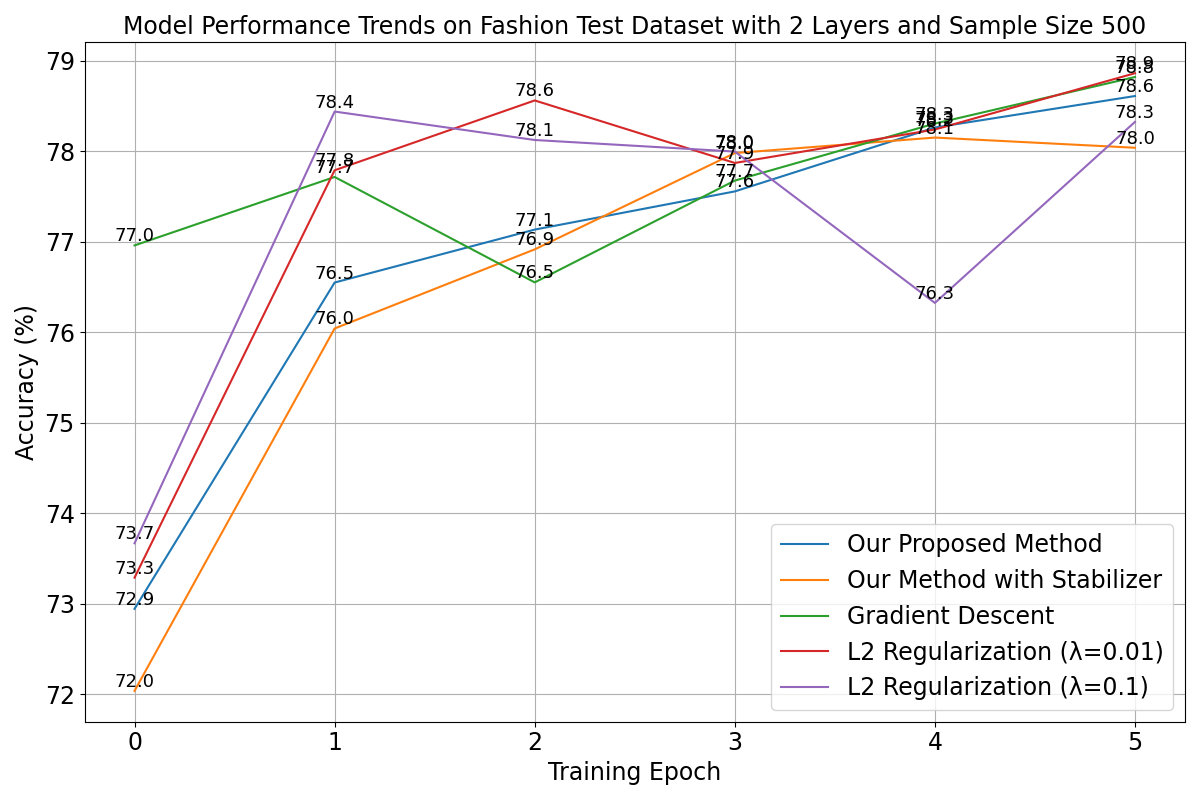}
\label{fig:Fashion_layers2_samples500}
\end{minipage}
\begin{minipage}{0.33\linewidth}
\centering
\includegraphics[width=0.95\linewidth]{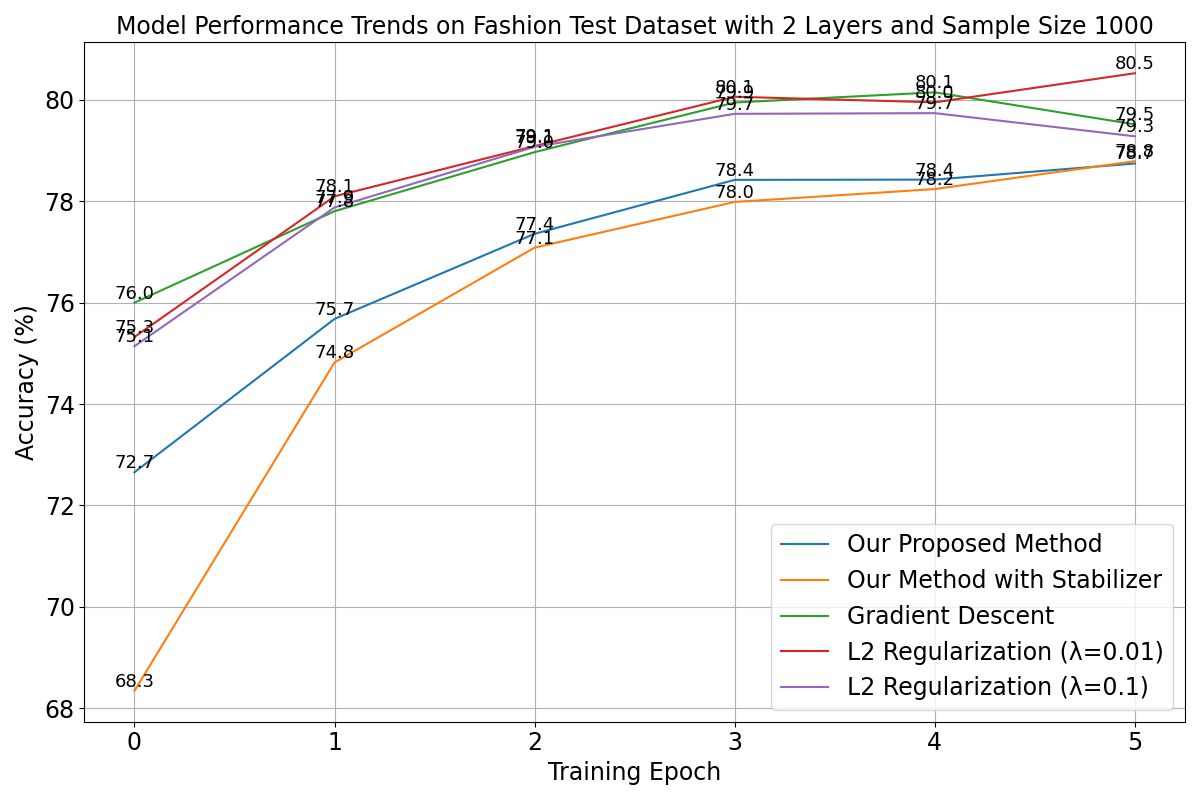}
\label{fig:Fashion_layers2_samples1000}
\end{minipage}
\\\\
\begin{minipage}{0.33\linewidth}
\centering
\includegraphics[width=0.95\linewidth]{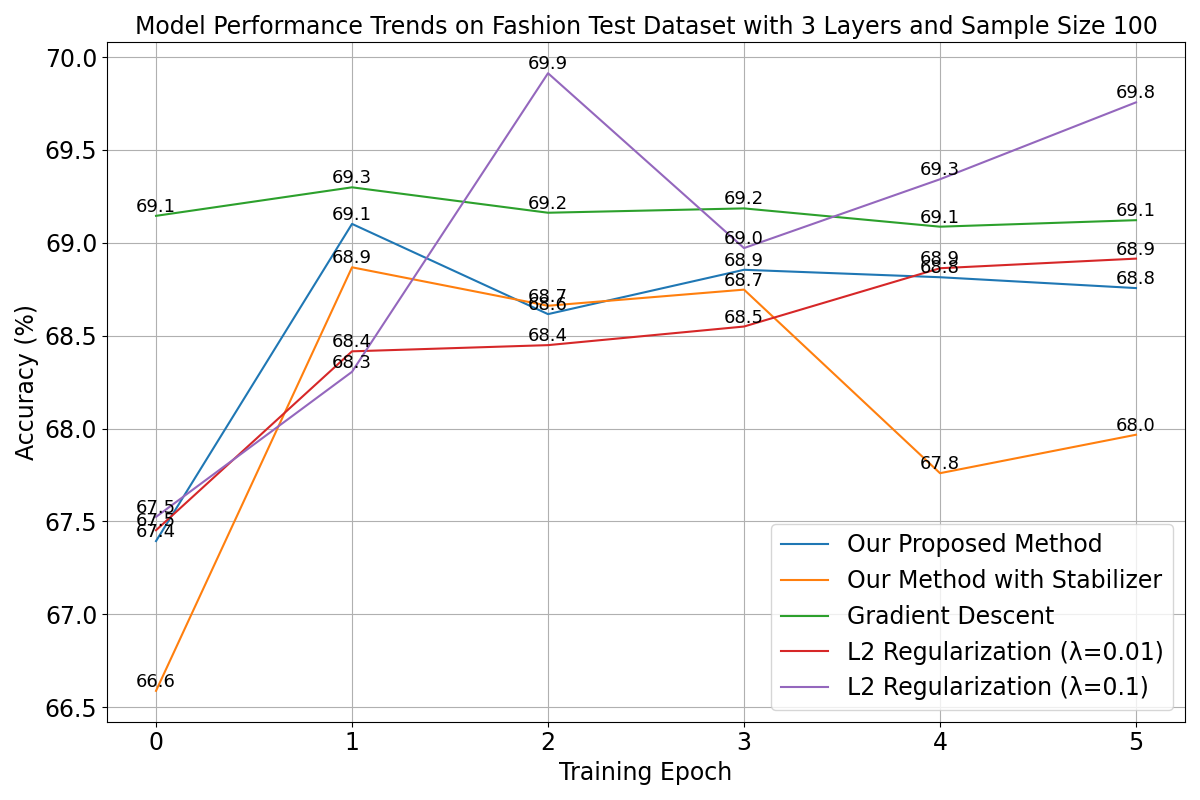}
\label{fig:Fashion_layers3_samples100}
\end{minipage}
\begin{minipage}{0.33\linewidth}
\centering
\includegraphics[width=0.95\linewidth]{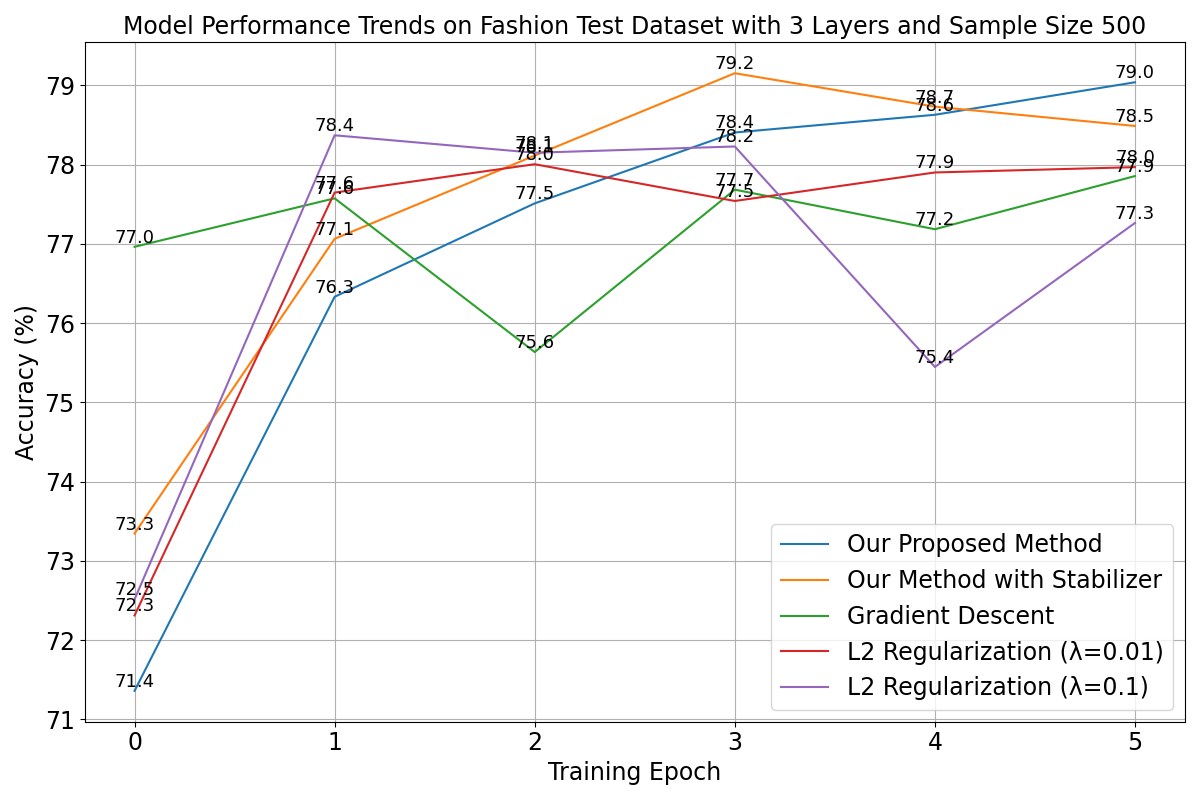}
\label{fig:Fashion_layers3_samples500}
\end{minipage}
\begin{minipage}{0.33\linewidth}
\centering
\includegraphics[width=0.95\linewidth]{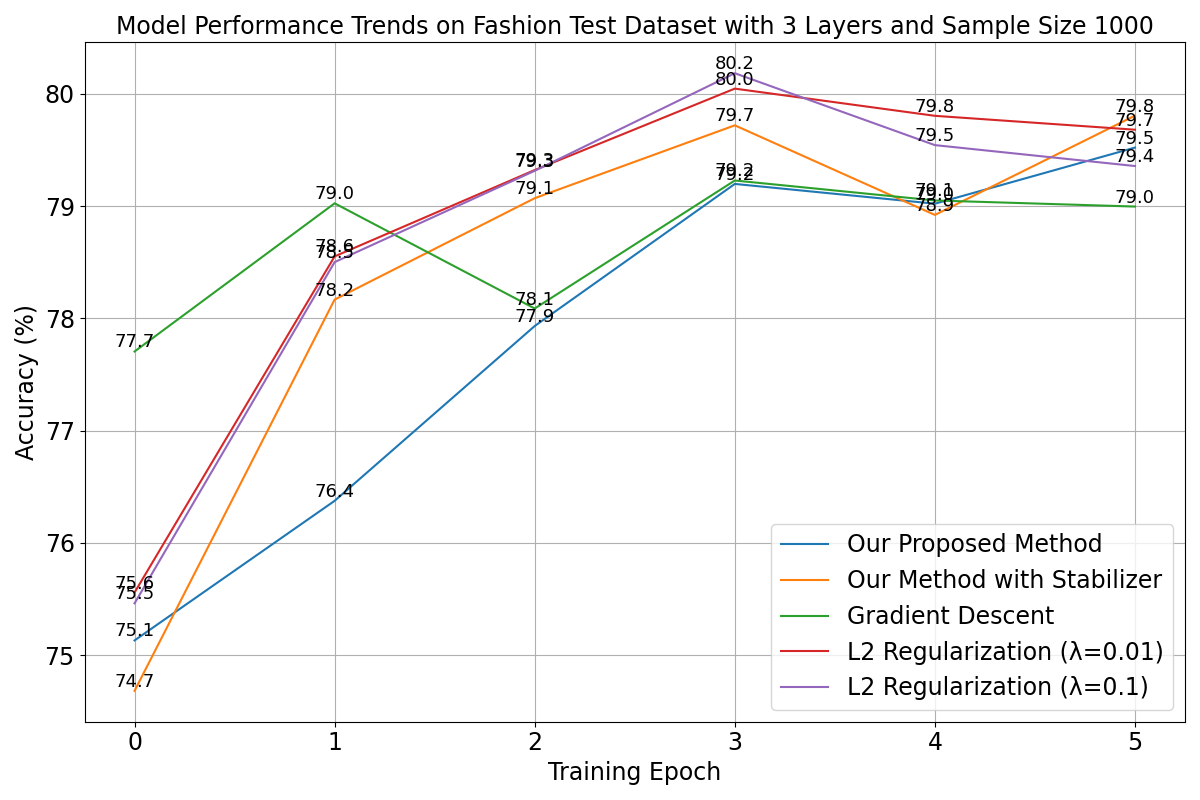}
\label{fig:Fashion_layers3_samples1000}
\end{minipage}
\caption{Results on Fashion dataset with models with 2 layers (Top row) and 3 layers (Bottom row). From left to right: Sample Size 100, 500, 1000.}
\end{figure}

\begin{figure}[h!]
\begin{minipage}{0.33\linewidth}
\centering
\includegraphics[width=0.95\linewidth]{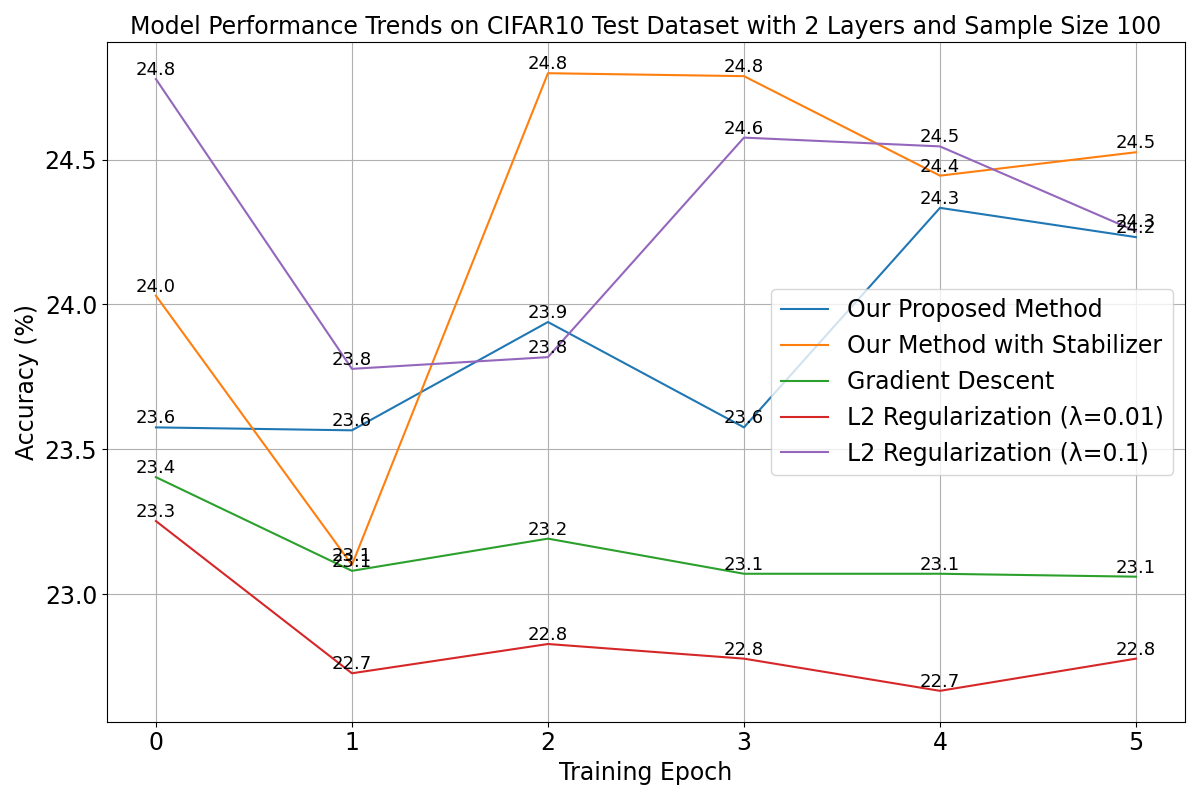}
\label{fig:CIFAR10_layers2_samples100}
\end{minipage}
\begin{minipage}{0.33\linewidth}
\centering
\includegraphics[width=0.95\linewidth]{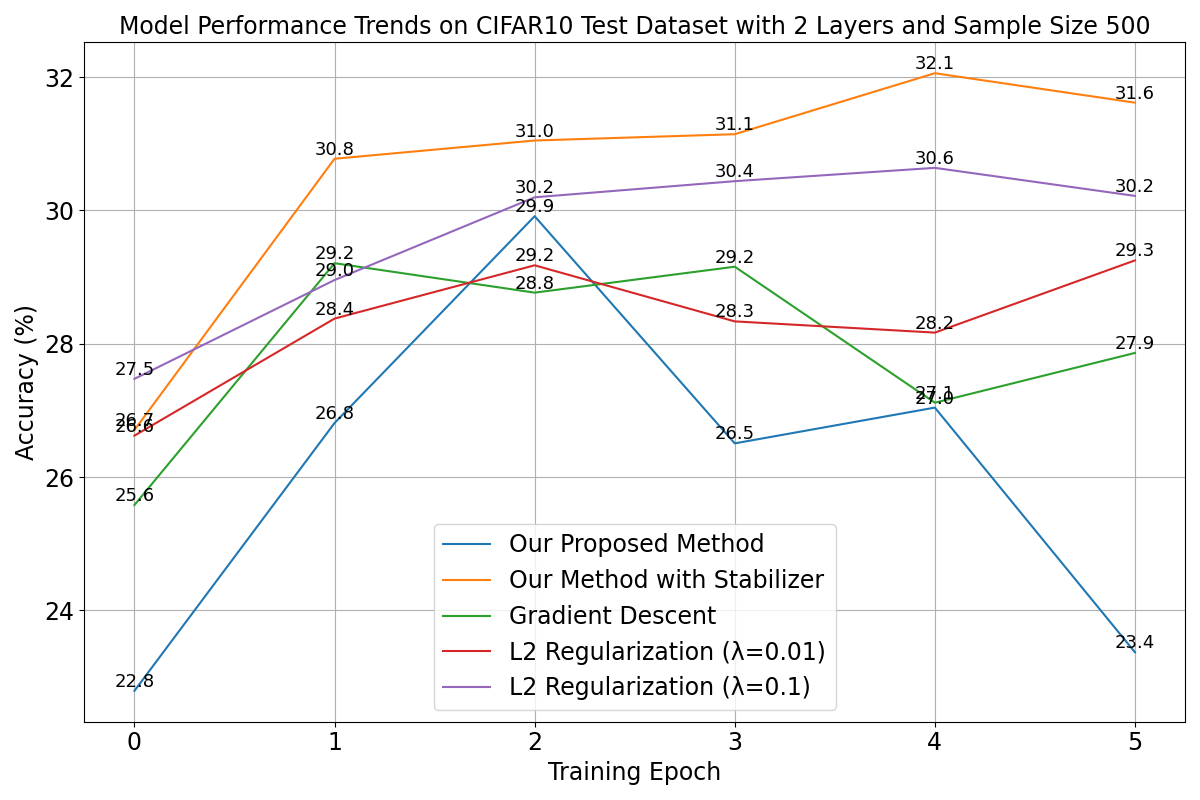}
\label{fig:CIFAR10_layers2_samples500}
\end{minipage}
\begin{minipage}{0.33\linewidth}
\centering
\includegraphics[width=0.95\linewidth]{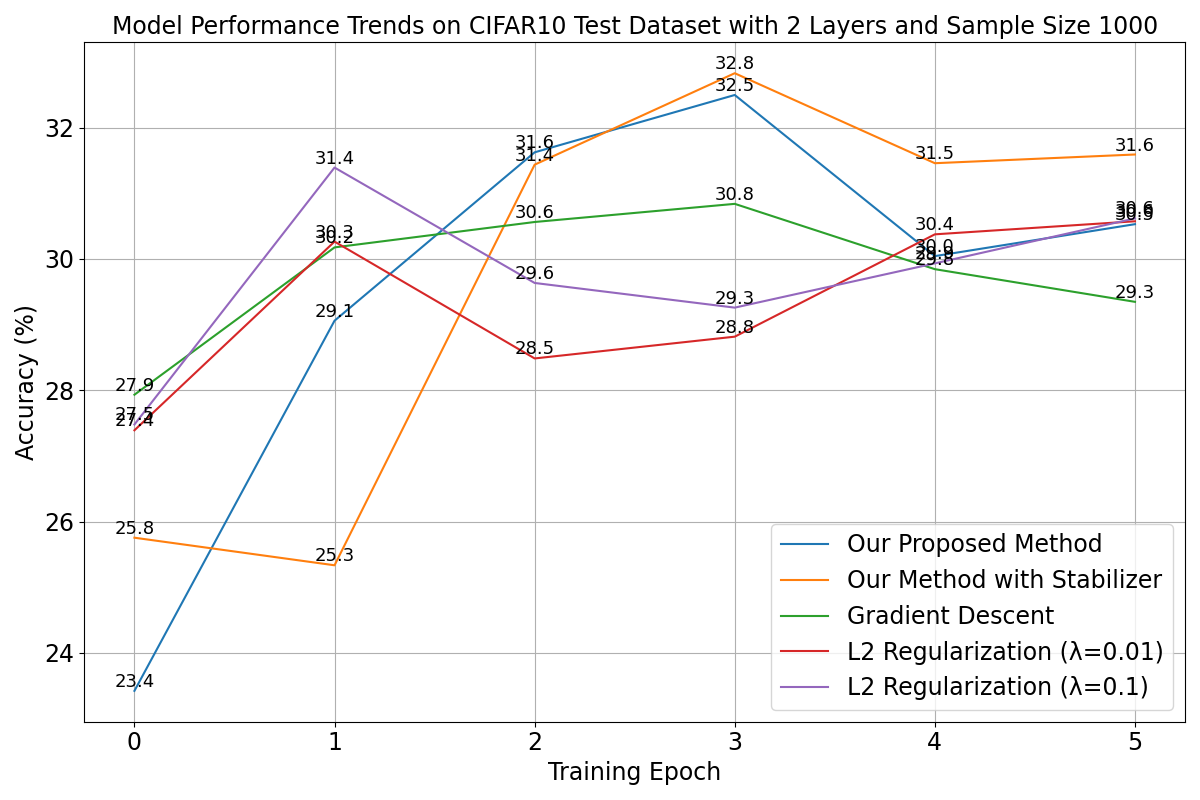}
\label{fig:CIFAR10_layers2_samples1000}
\end{minipage}
\\\\
\begin{minipage}{0.33\linewidth}
\centering
\includegraphics[width=0.95\linewidth]{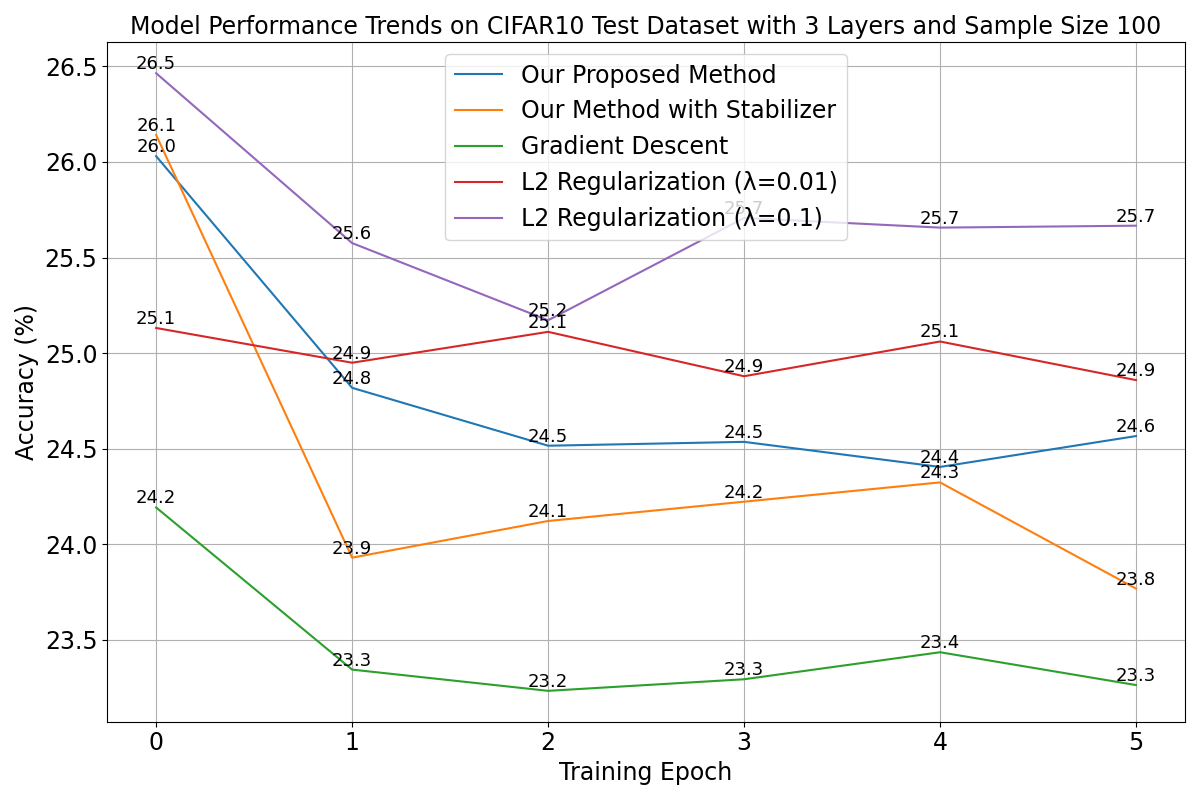}
\label{fig:CIFAR10_layers3_samples100}
\end{minipage}
\begin{minipage}{0.33\linewidth}
\centering
\includegraphics[width=0.95\linewidth]{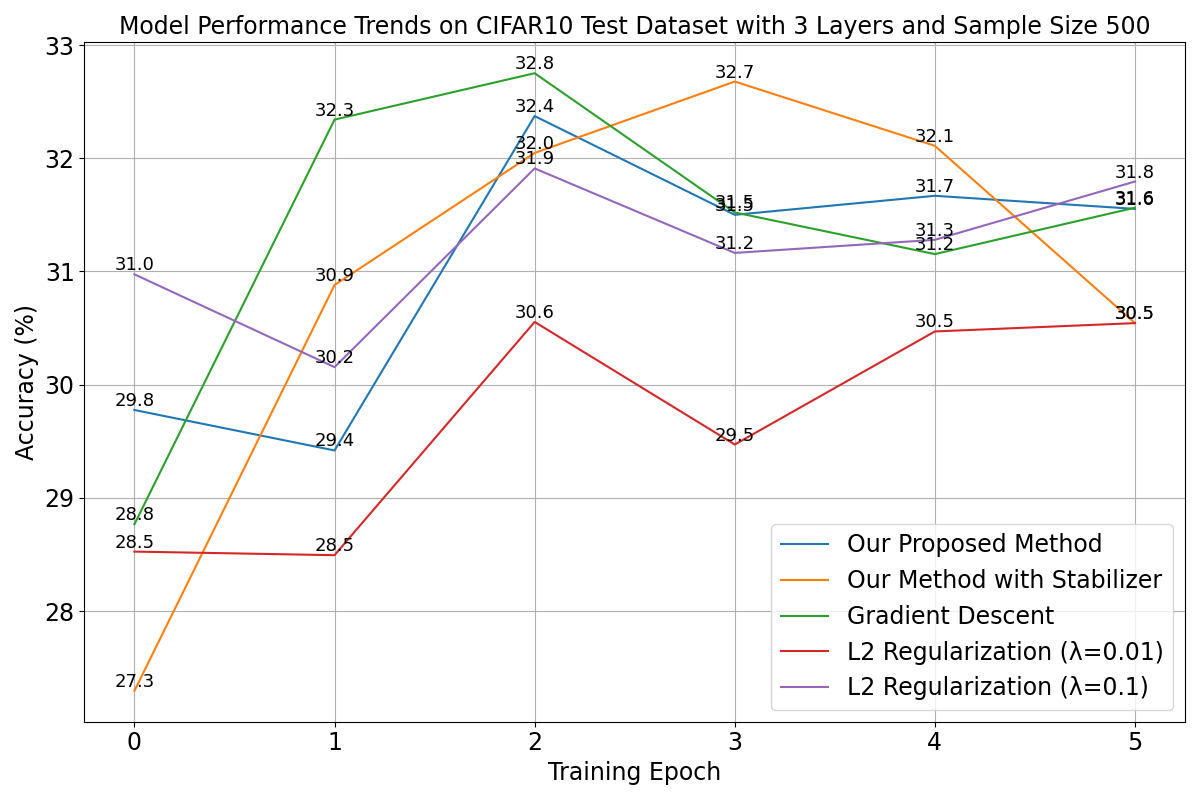}
\label{fig:CIFAR10_layers3_samples500}
\end{minipage}
\begin{minipage}{0.33\linewidth}
\centering
\includegraphics[width=0.95\linewidth]{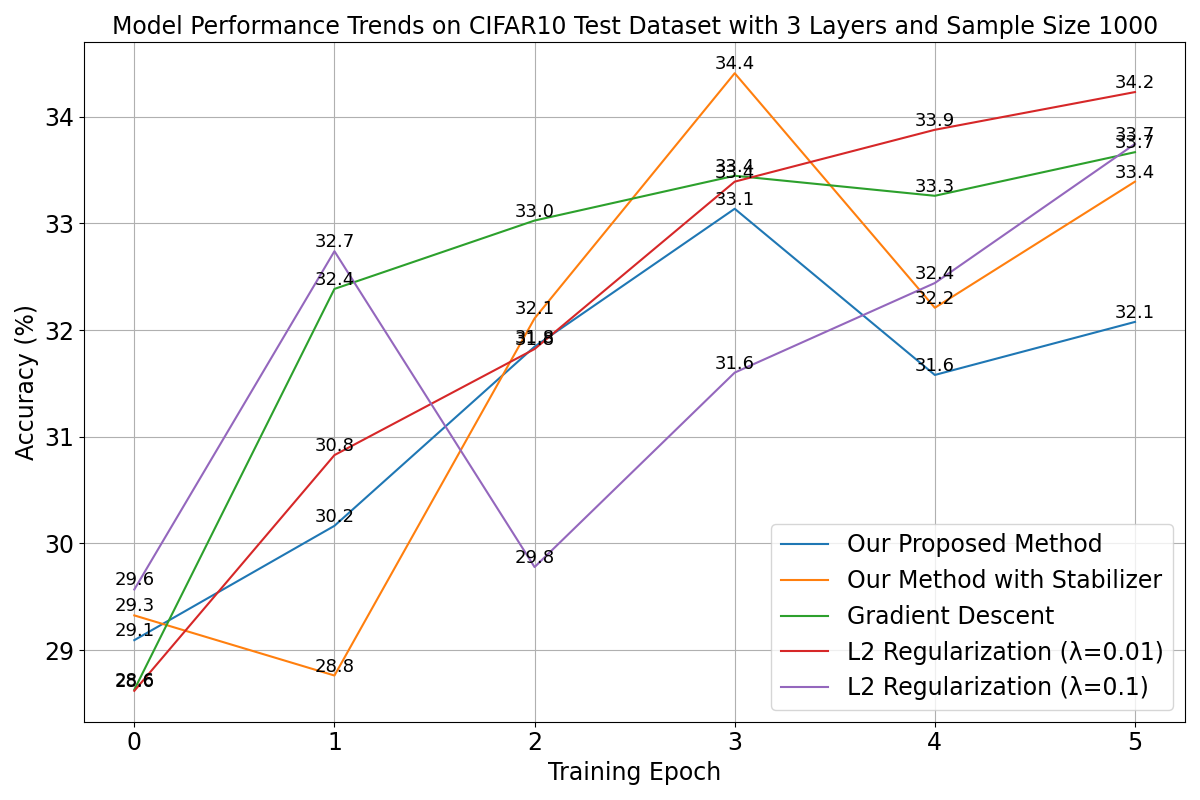}
\label{fig:CIFAR10_layers3_samples1000}
\end{minipage}
\caption{Results on CIFAR10 dataset with models with 2 layers (Top row) and 3 layers (Bottom row). From left to right: Sample Size 100, 500, 1000.}
\label{fig:CIFAR10_results}
\end{figure}

\end{landscape}
\end{document}